\documentclass[journal]{IEEEtran}
\usepackage{graphicx}
\usepackage{amsmath,amsfonts,amssymb}
\usepackage{capt-of,etoolbox}
\usepackage{amsfonts}
\usepackage{listings}
\usepackage{xspace}
\usepackage{multirow}
\usepackage{array}
\usepackage[table,xcdraw]{xcolor}
\usepackage{url}
\usepackage[draft]{hyperref}
\usepackage{mathtools}
\usepackage[font=small,labelfont=bf]{caption}
\usepackage{microtype}

\DeclareGraphicsExtensions{.pdf,.png,.jpg}

\newcommand{\algoname}{ARS}
\newcommand{\algonameext}{Augmented Reality Semiautomatic-labeling}
\newcommand{\elettrodataset}{Industrial}
\newcommand{\fruitdataset}{Fruits}
\newcommand{\histoname}{Viewpoint Coverage}

\newcommand{\eg}{\textit{e.g. }}
\newcommand{\ie}{\textit{i.e. }}

\newcommand{\norm}[1]{\left\Vert#1\right\Vert}

\newcommand{\TT}{\mathbf T}
\newcommand{\RR}{\mathbf R}
\newcommand{\Real}{\mathbb R}
\newcommand{\PositiveIntegers}{\mathbb Z^{+}}
\newcommand{\NInst}{V}

\ifCLASSINFOpdf
\else
\fi

\begin{document}

\makeatletter
\newenvironment{ntp}{%
  \@beginparpenalty\@lowpenalty
  \bfseries\small\textit{Note to Practitioners}\textemdash
  \@endparpenalty\@M}%
{\par}
\makeatother

\title{Semi-Automatic Labeling for Deep Learning in Robotics
\thanks{Daniele De Gregorio (d.degregorio@unibo.it), Alessio Tonioni (alessio.tonioni@unibo.it) and Luigi Di Stefano (luigi.distefano@unibo.it) were with the DISI department, Universit\`a degli Studi di Bologna, 40136 Bologna, Italy.}
\thanks{Gianluca Palli (gianluca.palli@unibo.it) was with the DEI department, Universit\`a degli Studi di Bologna, 40136 Bologna, Italy.}
}


\author{Daniele De Gregorio,  Alessio Tonioni, Gianluca Palli  and Luigi Di Stefano}



\maketitle
\begin{abstract}

In this paper, we propose Augmented Reality Semi-automatic labeling (ARS), a semi-automatic method which leverages on moving a 2D camera by means of a robot, proving precise camera tracking, and an augmented reality pen to define initial object bounding box, to create large labeled datasets with minimal human intervention. By removing the burden of generating annotated data from humans, we make the Deep Learning technique applied to computer vision, that typically requires very large datasets, truly automated and reliable. With the ARS pipeline we created effortlessly two novel datasets, one on electromechanical components (industrial scenario) and one on fruits (daily-living scenario), and trained robustly two state-of-the-art object detectors, based on convolutional neural networks, such as YOLO and SSD. With respect to conventional manual annotation of 1000 frames that takes us slightly more than 10 hours, the proposed approach based on ARS allows to annotate 9 sequences of about 35000 frames in less than one hour, with a gain factor of about 450. Moreover, both the precision and recall of object detection is increased by about 15\% with respect to manual labelling. All our software is available as a ROS package in a public repository alongside with the novel annotated datasets.

\end{abstract}

\begin{ntp}
This paper was motivated by the lack of a simple and effective solution for the generation of datasets usable to train a data-driven model, such as a modern Deep Neural Network, so as to make them accessible in an industrial environment.
Specifically, a deep learning robot guidance vision system would require such a large amount of manually labeled images that it would be too expensive and impractical for a real use case, where system reconfigurability is a fundamental requirement.
With our system, on the other hand, especially in the field of industrial robotics, the cost of image labelling can be reduced, for the first time, to nearly zero, thus paving the way for self-reconfiguring systems with very high performance (as demonstrated by our experimental results). 
One of the limitations of this approach is the need to use a manual method for the detection of objects of interest in the preliminary stages of the pipeline (Augmented Reality Pen or Graphical Interface). A feasible extension, related to the field of collaborative robotics, could be to exploit the robot itself, manually moved by the user, even for this preliminary stage, so as to eliminate any source of inaccuracy.
\end{ntp}

\begin{figure*}
 \includegraphics[width=1\linewidth]{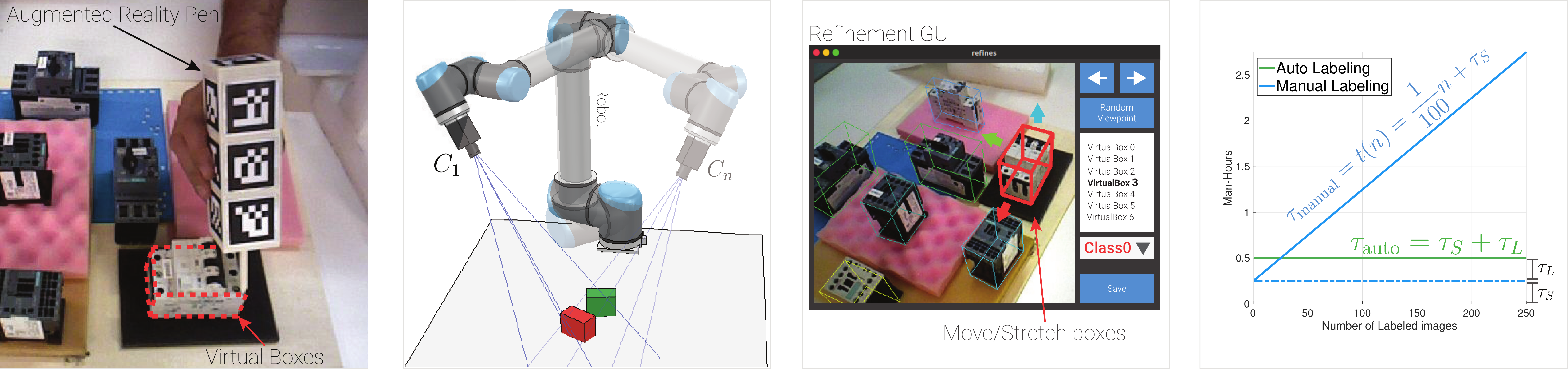}
\captionof{figure}{The system pipeline: 1) Draw virtual boxes around target objects; 2) Perform a scan acquiring camera poses (e.g. by a robot); 3) Refine virtual boxes; 4) Generate a training dataset. Therefore, with this approach it is possible to generate seamlessly an arbitrary number of self-labeled images without any human intervention. The right-most graph compares the time for manual labeling $\tau_\text{manual}$ (experimentally $\sim$100 images/hour) to that required by the proposed approach $\tau_\text{auto}$. The setup time $\tau_\text{S}$ is shared between the two modalities, $\tau_\text{L}$, instead, is the time required to label with the Augmented Reality Pen.
}
\label{fig:teaser}
\end{figure*}

\section{Introduction}
Many applications in modern robotics require a rich perception of the environment. The automatic detection of  known objects, for instance, may help in scenarios like quality control, automated assembly, pick-and-place, bin-picking and so on. The specific task of \emph{object detection}, as many others in computer vision, has witnessed dramatic advances in recent years due to the introduction of deep learning methods  capable of recognizing many different object categories in real time \cite{redmon2016yolo9000,huang2016speed,Lin_2017_ICCV}. Yet, these state-of-the-art methods mandate training on large datasets of images annotated with bounding boxes surrounding the objects of interest. The process of manually annotating the training images is time-consuming, tedious and prone to errors that may often lead to flimsy datasets or noisy annotations, especially when these are performed by non-professional users. Although a number of high-quality, large, multi-category training datasets are publicly available \cite{lin2014microsoft,Rennie2016,Calli2017}, they usually concern general classes (such e.g. person, car, cat, etc.) that may not suit the needs of a specific -- industrial -- task. In particular, robotic applications typically require detection of a relatively small set of specific object instances in cluttered and heavily occluded scenes captured from many different viewpoints, and with the sought objects possibly changing frequently overtime. In these settings, the richness and  fickleness of the training dataset play a fundamental role to any deep learning solution. Indeed, handling a new object may easily require thousands of annotated images, which translates into tons of man-hours.


As at such large scale manual annotation turns out impractical and often inaccurate, we propose a user-friendly approach that allows to gather effortlessly and almost automatically huge datasets of annotated images in order to train state-of-the-art 2D Object Detectors based on deep learning like \cite{redmon2016yolo9000,huang2016speed,Lin_2017_ICCV}, or even 3D Pose Estimators like \cite{Kehl_2017_ICCV}, \cite{xiang2017posecnn}, \cite{rad2017bb8}.
We start by acquiring a sequence of 2D images alongside with the camera pose in each frame. Based on this input information, our method deploys Augmented Reality techniques to enable the user to create easily the manual annotations for the first frame (or few ones) and then can deliver automatically accurate annotations for all the other frames of the sequence without any further human intervention. Hence, we dub our proposal  \textbf{\algoname{}: \algonameext{}}.  It is worth pointing out that the camera poses required by our method may be obtained in different manners, such as, e.g., by a monocular SLAM algorithm like \cite{mur2017orb,engel2014lsd} or a motion capture system. Yet, as in this paper we mainly address robotic applications, we propose leveraging on a camera directly mounted on a robotic arm in an \emph{eye-on-hand} configuration in order to gather tracked image sequences with high tracking accuracy, as also shown, \eg, in \cite{Zeng2016,de2016robotfusion,Mitash}. We expect such an approach to become increasingly more practical and affordable with the advent of lightweight,  but precise, \textit{collaborative robots}, allowing the dataset creation directly on the application scenario,  reducing in this way the gap, with regard to data distribution, between training and real conditions. 
We rely only on a plain 2D camera, which is cheap and does not set forth any restriction as long as objects are visible whereas a Stereo or RGB-D  camera would have hindered flexibility, e.g. due to constraints on the minimum object-camera distance (and thus object size) or the inability to sense poorly textured or black surfaces. We developed a publicly available ROS package
\footnote{\url{https://github.com/m4nh/ars}}
implementing all the tools described in this paper, which can be used to realize  the \algoname{} labelling pipeline starting from a video sequence with associated camera poses. Furthermore, we distribute all the datasets used throughout the experiments (see \autoref{sec:experimental}), which enables reproducibility of the experimental results. 

\section{Related work}\label{sec:related}
A popular approach to speed-up creation of training datasets consists in the use of synthetic images rendered \cite{mayer2016large,Ros2016,movshovitz2016useful,Carlucci2016} or even grabbed from realistic videogames \cite{Richter2016,Johnson-Roberson2016}. These techniques can deliver countless perfectly annotated images with  human effort/time spent only to build synthetic scenes. 
However, obtaining a large dataset of photo realistic images usually comes at a cost as it may require hours of highly specialized human work to construct suitable synthetic environments plus many hours of computation on high-performance graphical hardware for rendering. In some practical settings, useful synthetic objects may be available beforehand in the form of CAD models although, indeed quite often, the textures may either be missing or look quite diverse with respect to the appearance of the actual objects.  

Moreover, it is well known  \cite{movshovitz2016useful,Carlucci2016} that training deep neural network by synthetic images does not yield satisfactory performance upon  testing on real data due to the inherent difference between the ideal and real images, an issue often referred to as \emph{domain gap}  and calling for specific \emph{domain adaptation} techniques, such as, e.g., fine-tuning the network by  -- fewer -- manually  annotated real images. 
Recent works \cite{shrivastava2016learning,zhang2018fully,tzeng2017adversarial,bousmalis2017using} focus on  developing ad-hoc adaptation techniques to close the performance gap between training and test distribution. Unfortunately the performance achievable are still quite far from those obtainable training on real data or fine tuning on few annotated samples.
Alternatively, to ameliorate the domain shift, \cite{Georgakis2017} proposes an hybrid approach whereby an object detection system is trained on rendered views of synthetic 3D objects superimposed on real images; however, the blend between synthetic and real is far from perfect such that an additional fine-tuning  on a real dataset is still needed. 
Differently, in this paper we propose a methodology to ease and speed-up the acquisition of large labeled datasets of real images which may be acquired directly in the deployment scenario, thereby avoiding any gap between the training and test domains. 

Several approaches tailored to dataset creation have been proposed within the robotic research community:  \cite{Kendrick2017} proposes a system suitable for 3D face annotation, \cite{Zeng2016} proposes a semi-automatic technique to acquire a training dataset for object segmentation and \cite{Mitash} extends the idea to support object detection by leveraging on physical simulation to create realistic object arrangements. All this proposals require depth information from the sensor, and for the last two, realistic 3D models (\eg textured CAD models). A similar solution is proposed in \cite{nguyen2016robust} and \cite{wong2015smartannotator} where an environment is reconstructed by means of an RGB-D sensor and the labeling procedure is performed on the resulting 3D model. Conversely, our approach  does not need any clue about he 3D shape of the object nor does it require depth information at training or test time. Moreover, our approach is the first technique usable with very small objects (as shown in \cite{de2018integration}).

\section{Method description}\label{sec:method_description}

Given a set of images, each equipped with the 6-DoF pose of the camera, and knowing the pose of the observed objects \textit{w.r.t.} each vantage point, it is possible to project in each image some simplified representation of these items through augmented reality in order to generate automatically annotations (e.g. 2D bounding boxes, class labels, etc.) useful to train machine learning models.  \autoref{sec:input_dataset} describes formally the input data required by our proposed \algoname{} labeling pipeline, which, as depicted in \autoref{fig:teaser}, can be summarized in the following main steps:

\begin{enumerate}
\setcounter{enumi}{-1}
\item Scene Setup (\ie Arrange the objects randomly);\vspace{-0.0cm}
\item Outline virtual boxes around the target objects;\vspace{-0.0cm}
\item Scan the environment by a tracked camera;\vspace{-0.0cm}
\item Refine virtual boxes by visual analysis of the scan;\vspace{-0.0cm}
\item Generate automatically a training dataset.
\end{enumerate}

\noindent \algoname{}  can be used to generate a dataset starting from scratch (\ie following all the steps 0,1,2,3,4) or by exploiting archived material (\ie following only steps 3 and 4, assuming the availability of an off-line camera tracker algorithm, like \cite{mur2017orb}, to be applied to a recorded video sequence).
The reminder of this section will describe in detail all the important steps of the labeling pipeline: \autoref{sec:input_dataset} deals with the input data, \autoref{sec:augmente_pen} presents  a way to define virtual boxes by an Augmented Reality Pen directly interacting with the physical environment,  \autoref{sec:pose_refinement} addresses refining (or create a posteriori) virtual boxes around the objects and, finally, \autoref{sec:gen_training_data} describes the procedure used to generate annotated images. In \autoref{sec:notation}, we introduce the notation adopted throughout the rest of the paper.

\begin{figure*}
	\centering
	\begin{tabular}{ccc}
		\includegraphics[width=0.31\textwidth]{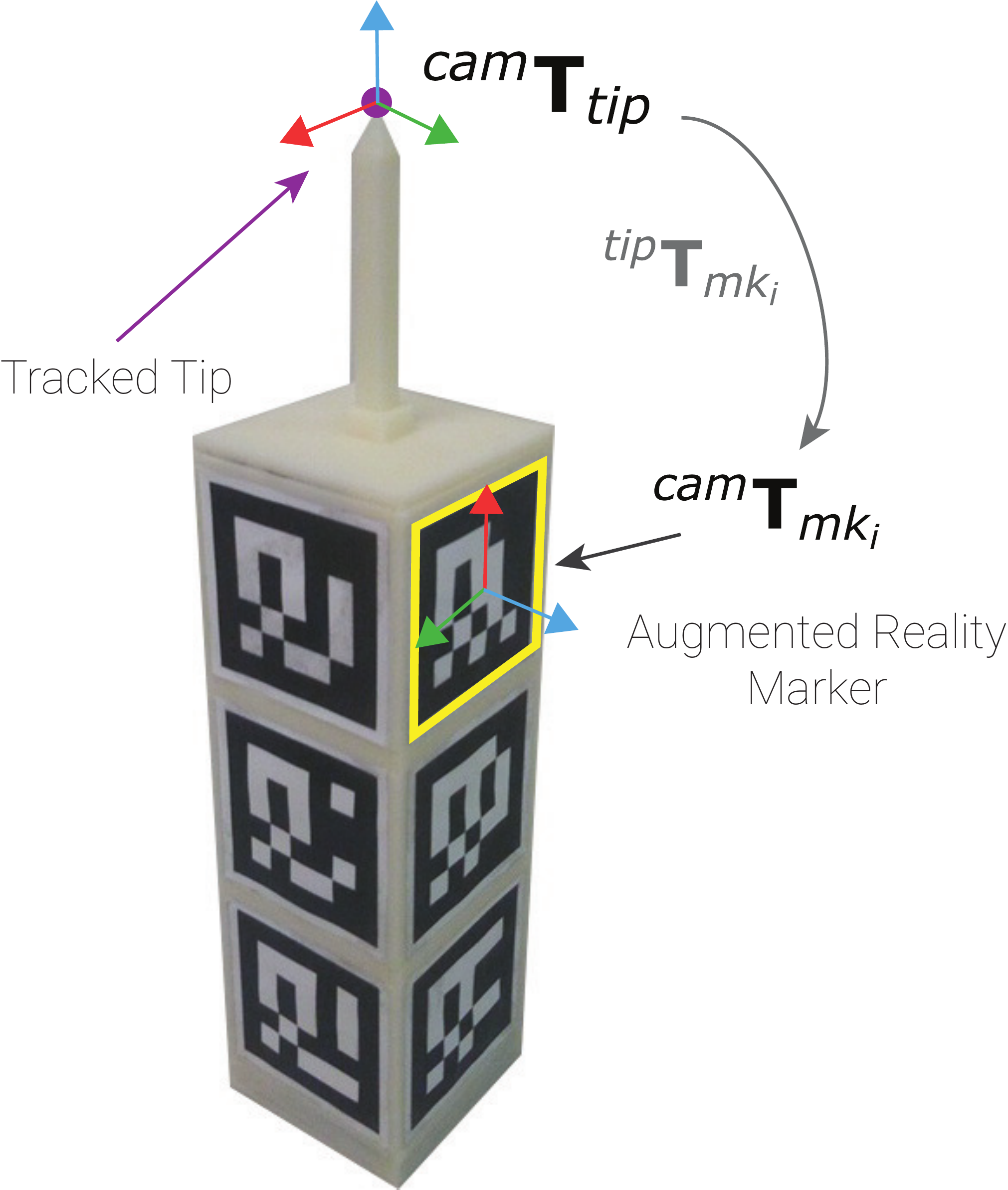}&\includegraphics[width=0.31\textwidth]{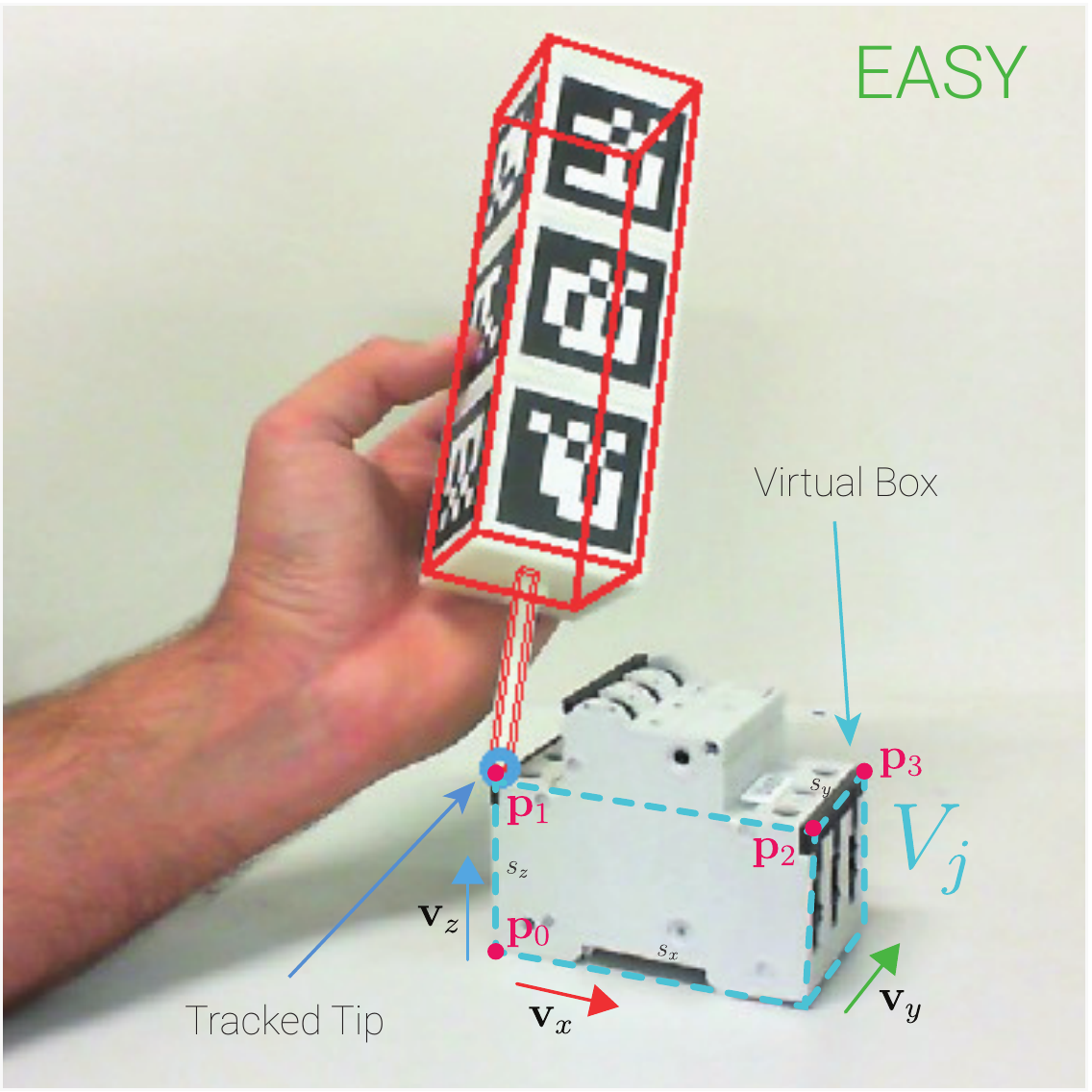}&\includegraphics[width=0.31\textwidth]{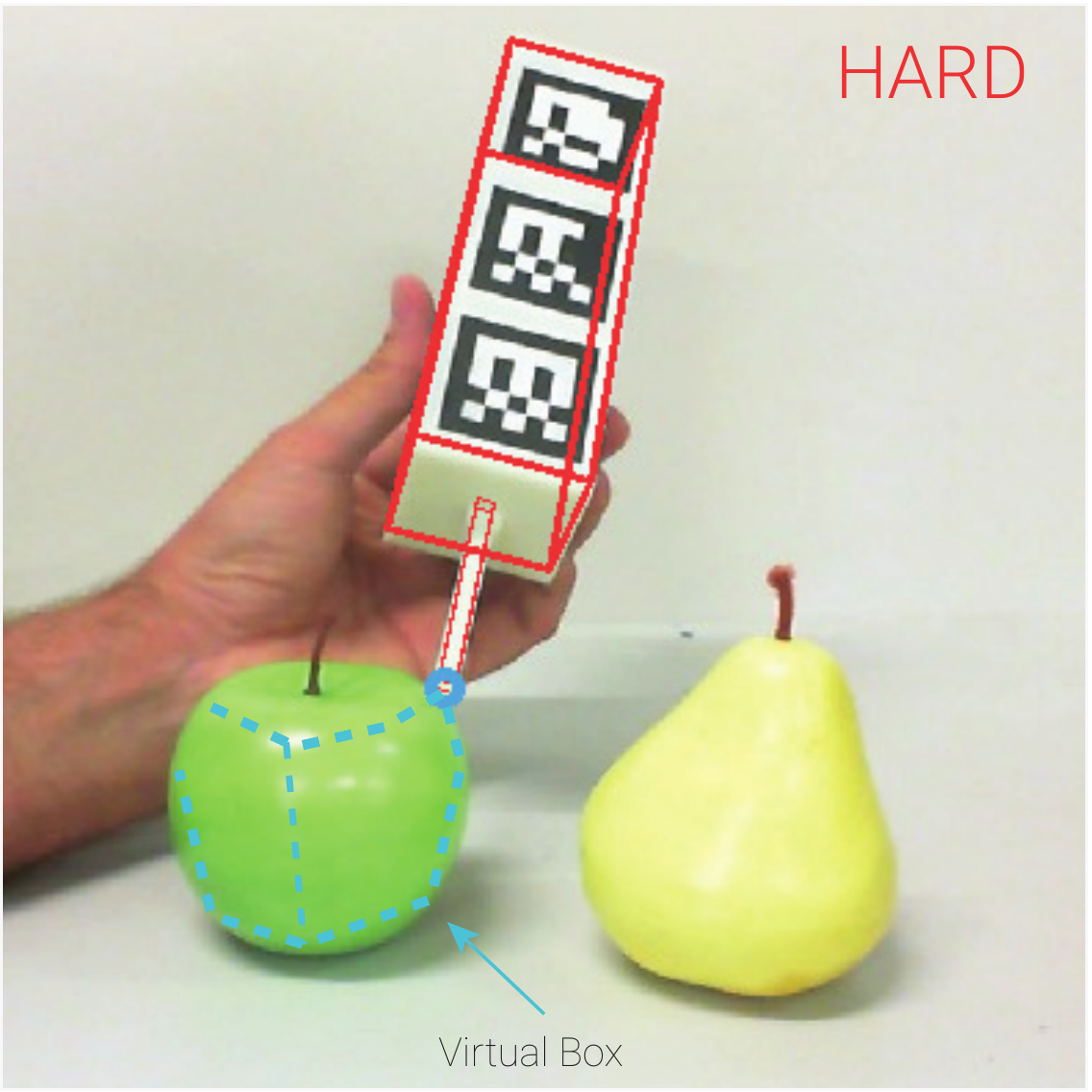}\\
		\textbf{(a)} ARP & 
		\textbf{(b)} Drawing on geometrical object & 
		\textbf{(c)} Drawing on rounded object \\
	\end{tabular}
	\caption{(a) The \emph{Augmented Reality Pen} (ARP) used to draw virtual boxes. The pen features several Augmented Reality Markers  with a known pose $^{tip}\TT_{mk_i}$ \textit{w.r.t.} the tip. (b) By tracking the tip position we can easily draw a virtual box around a target object by touching its edges. (c) Conversely, its not simple to draw a virtual box around a rounded object.}
	\label{fig:arp}
	
\end{figure*}

\subsection{Notation}\label{sec:notation}

We denote as  $^{A}\mathbf{T}_{B} \in \Real^{4 \times 4}$ a 3D reference frame (briefly RF) $B$ expressed in the base $A$. So  $^{0}\mathbf{T}_{cam}$ represents the RF linked to the camera in the $zero$ reference frame (\ie the \textit{world} RF).  $m_i$ denotes a generic image and  $b = \{ x_b,y_b,w_b,h_b, c_b \}$ a square region (box) therein, with  $(x_b,y_b)$ the coordinates of the center of $b$, $w_b$ and $h_b$ the \emph{width} and \emph{height}, respectively, and, optionally, $c_b \in \PositiveIntegers$ the \emph{class} of the object contained in $b$. 

\subsection{The input data}\label{sec:input_dataset}
The input data to our labelling pipeline consist of two separate sets $\mathbb{F}$ and $\mathbb{I}$:

\begin{equation}
\begin{gathered}
\mathbb{F} =  \{ F_i = \{^{0}C_i , m_i \} , i \in [0,...,n] \} \\
\mathbb{I} = \{ \NInst_j=\{^{0}\TT_j , s_j, c_j \} , j \in [0,...,k] \}
\end{gathered}
\end{equation}\label{eq:dataset}
 
\noindent $\mathbb{F}$ represent the acquired images (\emph{frames}) $F_i=\{^{0}C_i , m_i \}$ with $^{0}C_i$ being the camera matrix for image $m_i$. Each $^{0}C_i$ can be expressed as the multiplication of the \emph{intrinsics} matrix $\hat{A} \in \Real^{4\times4}$ encoding camera specific parameters and the \emph{extrinsics} matrix:

\begin{equation}
^{0}\TT_{cam_i} =\small{ \begin{bmatrix}
    \RR_{cam_i}  & \mathbf{t}_{cam_i} \\
    0 & 1
  \end{bmatrix}}\in \Real^{4\times4}
\end{equation}

\noindent encoding camera orientation ($\RR_{cam_i} \in \Real^{3\times3}$) and position ($\mathbf{t}_{cam_i} \in \Real^{3}$) with respect to the \textit{world} frame.


\noindent $\mathbb{F}$ can be obtained using any method to track the camera movement, for example: a motion capture system, a SLAM based method like \cite{mur2017orb} or a camera mounted on a robotic arm in an \textit{eye-on-hand} configuration.
The $\mathbb{I}$ set, instead, corresponds to a collection of $k$ object instances present in the scenes. Each instance can be thought of as a 3D Virtual box, as shown in \autoref{fig:arp}(b-c), and can be expressed  as a tuple $\NInst_j=\{^{0}\TT_j , \mathbf{s}_j, c_j \}$ with $^{0}\TT_j$ the 6-DoF pose of the instance in the \textit{world} reference frame, $\mathbf{s}_j \in \Real^{3}$ the box dimensions and $c_j \in \PositiveIntegers$ its \emph{class}. In the following we will cover different methodologies to collect a suitable $\mathbb{I}$ set.

\subsection{Online Labeling by the Augmented Reality Pen}\label{sec:augmente_pen}

To create $\mathbb{I}$ quickly and effortlessly, we developed the 3D printed artifact pictured in \autoref{fig:arp}, which resembles a pen covered with Augmented Reality Markers (in short: Markers \cite{munoz2012aruco}); we will refer to this device as to the \emph{ARP} (Augmented Reality Pen). This tool can be used for labeling \emph{online} object instances  by interacting directly with the environment.
Each Marker on the ARP has a known pose  $^{tip}\TT_{mk_i}$ \textit{w.r.t.} the tip of the pen (the placement is CAD-driven). Using the OpenCV Marker Detector\footnote{\url{https://opencv.org/}} and the calibrated camera parameters we can estimate the pose of each of the markers, and consequently of the tip, \textit{w.r.t.} the camera $^{cam}\TT_{mk_i}$, then : $\small ^{cam}\TT_{tip} = ^{cam}\TT_{mk_i} \cdot (^{tip}\TT_{mk_i})^{-1}$.

To achieve proper vertices estimation, the ARP is tracked exploiting multiple markers simultaneously, each of which contributes to refine the estimated tip position:
 by averaging those given by all the visible markers we can obtain a more accurate estimation of the real position.
Accuracy therefore depends very much on the type of marker used, and has therefore not been dealt over in detail in this paper. As an indication, using this type of squared marker results in an error on the pose estimation proportional to the angle of inclination of the marker itself. 
 A more detailed explanation of this approach, along with an accuracy analysis, is described in \cite{jiawei2010three} and \cite{wu2017dodecapen}.
In particular, in \cite{wu2017dodecapen} a dodecahedron (instead of our parallelepiped-shaped pen) is used to reduce the angle of view of visible markers, thus reducing the estimation error.
 As shown in \autoref{fig:arp}(b), as the ARP can be tracked while being used to -- virtually -- draw a box around an object, those 3D points can then be used to construct the corresponding tuple $\NInst_j = \{^{0}\TT_j , s_j, c_j \}$. Our approach  requires only four specific points $p_0,p_1,p_2,p_3$ placed as shown in \autoref{fig:arp}(b). Then, from those spatial positions we can obtain the corresponding components of $\NInst_j$ as:


\begin{equation}\label{eq:virtual_object_buildings}
\begin{gathered}
^{cam}\TT_j =  
\begin{bmatrix}
    \mathbf{v}_x & \mathbf{v}_z \times \mathbf{v}_x & \mathbf{v}_z  & \mathbf{p_0} \\
    0 & 0 & 0 &  1
  \end{bmatrix} 
  \\
  \\
\mathbf{v}_z =\frac{• \mathbf{p}_1 - \mathbf{p}_0}{\norm{ \mathbf{p}_1 - \mathbf{p}_0}}   ,
\mathbf{v}_x =\frac{• \mathbf{p}_2 - \mathbf{p}_1}{\norm{ \mathbf{p}_2 - \mathbf{p}_1}}
\\
\\
s_j = 
\begin{bmatrix} s_x \\ s_y \\ s_z \end{bmatrix} =
\begin{bmatrix}
\norm{ \mathbf{p}_2 - \mathbf{p}_1} \\
\norm{ \mathbf{p}_3 - \mathbf{p}_2} \\ 
\norm{ \mathbf{p}_1 - \mathbf{p}_0} 
\end{bmatrix}
\end{gathered}
\end{equation}

\noindent The class $c_j$ should instead be specified by the user. The  
\autoref{eq:virtual_object_buildings} specifies the RF of the virtual object $^{cam}\TT_j $ referred to the camera RF; however, it can be easily transformed into the \textit{world} RF by knowledge of the current camera pose $^{0}\TT_j = {^{0}}\TT_{cam}  \cdot {^{cam}}\TT_j $.


By exploiting the ARP, a tight 3D bounding box can be easily created around box-like objects, such as in the case of electromechanical components, see \autoref{fig:arp}(b). However, this method will not work properly in case of arbitrarily shaped objects, such as the fruits depicted in \autoref{fig:arp}(c). To overcome this limitation, an off-line procedure to sketch $V_j$ by labeling a pair of frames picturing the same object from different vantage points has been developed, as reported in the following.

\subsection{Offline Labeling}\label{sec:pose_refinement}

\begin{figure}[t]
	\centering
		\includegraphics[width=1\columnwidth]{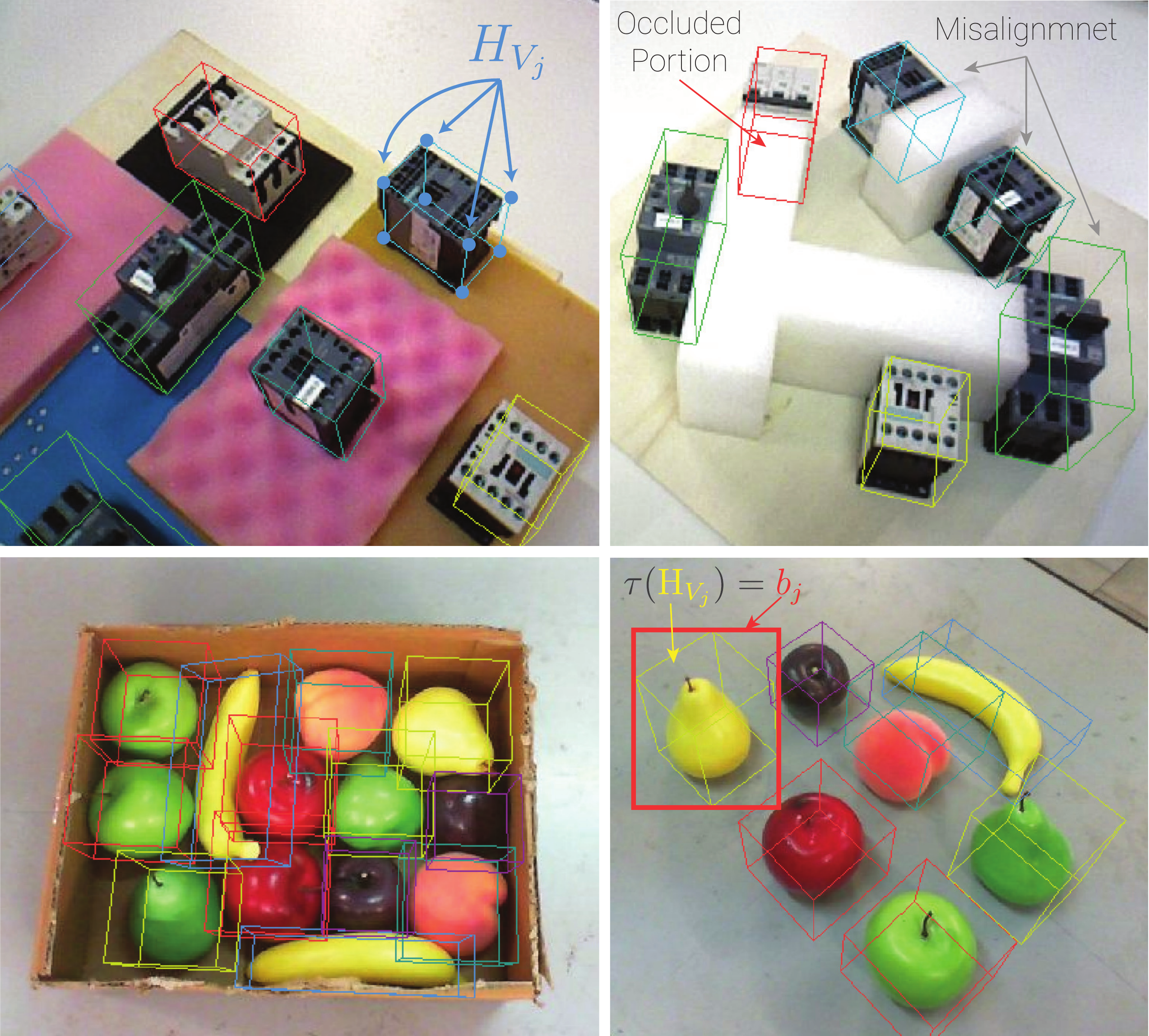}
	\caption{ Four random \emph{frames} acquired during the creation of our new dataset: the first row refers to the \elettrodataset{} dataset and displays virtual boxes drawn by the ARP; the second row shows samples from the \fruitdataset{} dataset with annotations created off-line by the technique described in \autoref{sec:pose_refinement}.}
	\label{fig:refinement_screens}
	
\end{figure}

\begin{figure}
	\centering
		\includegraphics[width=1\columnwidth]{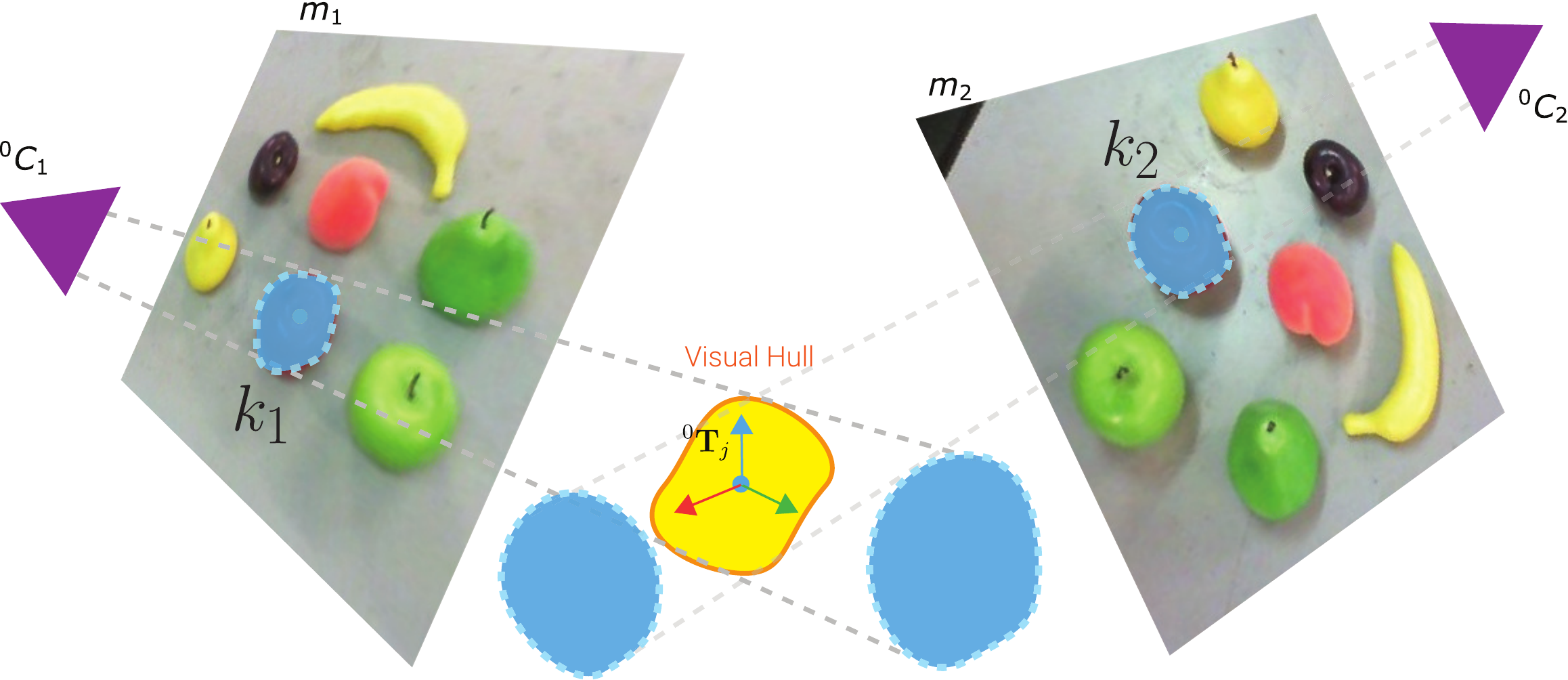}
	\caption{Graphical representation of the Visual Hull procedure used to build a virtual object from multiple 2D images.}
	\label{fig:frustum}
	
\end{figure}

In case of object with non-boxed shape, an offline procedure has been developed to increase the annotation accuracy with respect to what can be achieved by the ARP method previously described. The Fruits Dataset has been built by exploiting this off-line approach with good results, as shown in \autoref{fig:refinement_screens}. The offline procedure is based on a graphic interface through which the user can manually draw suitable masks around at least two different views of the object directly on the image frame. The complete procedure is detailed in the following.

As shown in \autoref{fig:refinement_screens}, upon acquisition of $\mathbb{F}$ and $\mathbb{I}$ it is quite straightforward to display frames from $\mathbb{F}$ with superimposed a 2D re-projection of the object instances defined in $\mathbb{I}$: each virtual object $\NInst_j$ can be represented as a list of 3D points corresponding, in this specific case, to the eight vertices of the box. By arranging the vertices as columns of the matrix $^{0}P_{V_j} \in \Real^{4 \times 8}$ ($4$ rows are a result of the homogeneous coordinates conversion) and converting them in the $i_{\text{th}}$ camera RF, $^{cam}P_{V_j}=^{cam_i}\TT_{0} \cdot {^{0}}P_{V_j}$, we can simply perform a 3D-to-2D re projection through:

\begin{equation}
\begin{bmatrix}
\lambda  H_{V_j} & 1
\end{bmatrix}^{\intercal}  =  \hat{A} \cdot {^{cam}}P_{V_j}  
\end{equation}

\noindent where $H_{V_j}$ is the set of corresponding 2D points and $\lambda$ the scale factor.
This procedure is quite general and can be applied to any set of points (\eg instead of virtual boxes we could have used virtual squares made by only 4 points if we are dealing with planar objects or arbitrary complex polygons for arbitrary shaped objects).
 \autoref{fig:refinement_screens} shows many examples of reprojected virtual boxes $H_{V_j}$ each one being the 2D re projection of the virtual box $V_j$. 
However, as shown in \autoref{fig:refinement_screens}(top-right), the $H_{V_j}$ produced with the ARP tool can sometimes not be highly accurate due to several nuisances (\eg the user hand-shake); therefore, we included in our software GUI a graphical tool that shows in augmented reality both the object and its bounding box from multiple points of view, as if the user was immersed in a 3D environment (see the third frame of \autoref{fig:teaser}), by offering a browsable \textit{Mixed Reality} scene \cite{milgram1994taxonomy}.

This tool allows also to manually edit the bounding boxes, in order to correct the position, orientation and size of the virtual objects. Moreover, through the same GUI also a novel technique can be exploited to annotate the fruit dataset, in order to estimate the position of an object by analyzing two tracked frames (i.e. two rgb images of which I know the exact camera pose) without any initial guess.
 The aforementioned method consists simply in estimating a 3D virtual box (or a more complex 3D shape) by manually drawing its 2D re-projection on at least two frames, as depicted in \autoref{fig:frustum}.
Considering a pair of frames $F_1,F_2$ we can draw two 2D masks $k_1,k_2$ around the appointed object (the red apple in this case); knowing the two camera matrices $^{0}C_1,^{0}C_2$ we can compute two view frustums which shall intersect -- likely -- in the center of mass of the real object producing what is generally called \emph{Visual Hull} (VH) \cite{baker2005shape}. The pose of the new $V_j$ (\ie $^{0}\TT_j$)  for the translational part can be computed easily using the center of mass of the VH, while for the rotational part we can choose the canonical identity matrix $I \in \Real^{3 \times 3}$ as a starting point. Finally, to complete the new $V_j$, we use the minimum bounding box algorithm, over the produced VH, to compute its dimensions $s_j$. It is important to note that a VH tends to be equal to the real 3D shape of the target object with the increase of the input frames labeled with the abovementioned technique, however the coarse $V_j$ created by this pipeline can be always manually refined using the already described interactive procedure.

\subsection{Genaration of the Training Data}\label{sec:gen_training_data}

The final step of the \algoname{} labeling pipeline, once both $\mathbb{F},  \mathbb{I}$ have been acquired and refined, is the creation of a training dataset suitable for modern machine learning models. 
In this work, as stated in the introduction, we mainly address the object detection task, which we will consider as an explanatory use case to show how to generate a training set. Our goal is to create a dataset consisting of images annotated with labeled 2D bounding boxes, $b_j$, surrounding each of the visible target objects. This information can be attained straightforwardly by simply reprojecting in each frame of $\mathbb{F}$ the 3D virtual objects in $\mathbb{I}$, \ie $H_{V_j}$, and then computing a function to produce a squared bounding box $b_j$ from it. A graphical representation of the process is depicted in   \autoref{fig:refinement_screens} (bottom-right), where it is clear that $b_j = \tau(H_{V_j})$ is obtained through function $\tau(\cdot)$ (\ie \textit{the minimum 2D bounding box}), which indeed may  be replaced by any custom function to obtain different kinds of labeled data (\eg \textit{the convex hull}).


\section{Experimental evaluation}
\label{sec:experimental}

\begin{figure}
	\centering
	\begin{tabular}{ccc}
		\includegraphics[width=0.45\columnwidth]{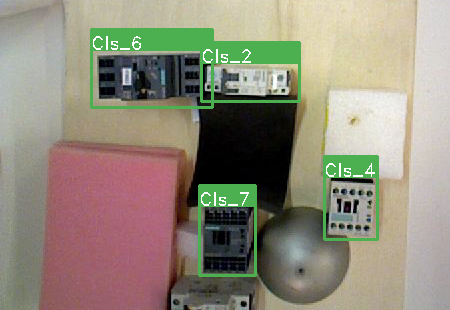}&\includegraphics[width=0.45\columnwidth]{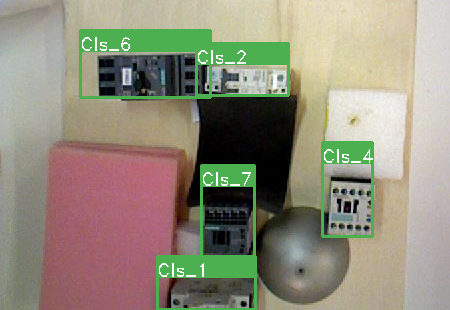}\\
		\textbf{(a)} Manual \elettrodataset & \textbf{(b)} Auto \elettrodataset 
	\end{tabular}
	\caption{Samples from the dataset used for our experiments. The green rectangles display the annotated bounding boxes, the white text over each box is the class of that box. (a) Electromechanical components dataset annotated manually, (b) Electromechanical components dataset with auto-generated labels.}
	\label{fig:dataset}
	
\end{figure}

To validate the \algoname{} pipeline we performed three sets of tests on two novel datasets that we are going to introduce in \autoref{ssec:datasets}. In \autoref{ssec:annotation}, we compare our automatic labeling procedure against a manual one. In \autoref{ssec:objDetTest}, we show how our datasets can be used to train CNN based object detectors. In \autoref{sec:slam_comparison}, we will compare monocular SLAM versus the use of a Robot for tracking the camera pose. Finally, in \autoref{ssec:histo} we introduce a new and interesting way of analyzing datasets, made possible only by our labeling approach.
Some qualitative results are shown in \autoref{fig:live_detections} as in the supplementary material, which also features a live demo of the \algoname{} labeling procedure from scratch.

\subsection{Datasets and evaluation metrics}
\label{ssec:datasets}
To validate our proposal we choose as test beds two different detection tasks: one concerning  recognition of 7 types of electromechanical components (\emph{\elettrodataset{}}), the other with 5 classes of fruits (\emph{\fruitdataset{}}); some samples from the two datasets are depicted in \autoref{fig:live_detections}. The first task concerns \textit{instance detection}, as the items display low intra and inter class variability: the same components are seen across all the images from different vantage points, with different components looking remarkably similar. This kind of instance detection is used in the WIRES project
(described in \cite{wires2018})
to implement a \textit{quality control} and \textit{automated assembly} systems for switchgear. The second task, instead, concerns \textit{object detection}, as items show high intra and inter class variability: each fruit is quite different from the others, yet also fruits belonging to the same class can show quite different appearances, e.g. in our acquisition we have two kinds of apples and pears showing different peel colors. The produced fruits object detector can be used for a simple \textit{pick\&place} manipulation application. The first dataset is composed of 9 acquisition ($\sim36000$ frames), the second by 8 shorter ones ($\sim7500$ frames). 

Both datasets were built by means of a camera mounted
 on an industrial manipulator, a COMAU Smart Six, with a position repeatability lesser than $0.05mm$. Using the industrial manipulator to move the camera we achieved a nearly perfect camera tracking, by computing the 6-DoF sensor pose through the robot kinematics, as to establish an upper bound for the labeling performance of \algoname{}. 
We will show in \autoref{sec:slam_comparison} how for less constrained applications, such us mobile robotics or home service robots, \algoname{} could rely on a classical monocular SLAM pipeline for tracking the camera (although sacrificing some accuracy).

We chose one sequence from \elettrodataset{} dataset and two from the \fruitdataset{} dataset, manually annotating them, to be used as test sets for the trained object detectors, we will refer to them as \elettrodataset{}\_Test and \fruitdataset{}\_Test, respectively. The other sequences are randomly rearranged to create sets with increasing number of samples, each splitted in 80\% train and 20\% validation. We will refer to each one of such set as: \texttt{$\langle$dataset name$\rangle$\_$\langle$number of samples$\rangle$}, \emph{e.g.} \elettrodataset{}\_1000 identifies 1000 sample from the training sequences of \elettrodataset{} that will be split into 800 training samples and 200 for validation. 
All datasets have been automatically annotated with \algoname{} and for further tests we enriched \elettrodataset{}\_1000 with manual annotation as well. We will use a "\_M" suffix for a manually annotated dataset (e.g. \elettrodataset{}\_1000\_M) and "\_A" suffix for annotation using \algoname{} (e.g. \elettrodataset{}\_1000\_A). 
\autoref{fig:teaser} shows a graphic comparison of the different efforts in human work hours needed to manually annotate a dataset (growing linearly with the number of required images) vs using \algoname{} (constant once sequences are acquired and virtual boxes are created). For reference, factoring out the common acquisition time, the manual annotation of 1000 frames took us slightly more than 10 hours, while using \algoname{} we were able to annotate all the 9 sequence of the \elettrodataset{} dataset in less than an hour ($\sim35000$ frames), with a gain of factor $\sim 450$.


To measure the detector performance we will use the standard object recognition metrics defined for the PASCAL VOC challenge \cite{Everingham10}. Given a prediction $b_j^p$ and the corresponding ground truth box $b_i$, we consider $b_j$ correct if they have the same class and ${IOU(b_j^p,b_i)>IOU_{th}}$ with $IOU(\cdot)$ intersection over union of the boxes and $IOU_{th}$ a threshold parameters. Given the set of correct predictions we can measure: \emph{Precision},\emph{Recall} and \emph{average intersection over union for correct predictions (avgIOU)}. Usually a detector produces quite a lot of $b_j^p$ each one associated with a certain confidence value $t_j^p \in [0,1]$, by thresholding the minimum confidence allowed we can tune the behaviour of the system. We represent the global performance with different confidence threshold using Precision/Recall curves and, concisely, with the \textbf{mean average precision}($mAP$), defined as the approximation of the area under the precision recall curve. 

\subsection{Annotation Study}
\label{ssec:annotation}

In this section we want to test if the annotation obtained by \algoname{} resemble what a human annotator would do. To verify this,  we have compared the two sets of annotations, manual and auto-generated, by considering the first as the output of an ideal detector, while the second as ground truth annotations.
Using $\small\text{IOU}_{th}=0.3$ we obtain the following performance: $\small\text{Precision}=98.49\%$, $\small\text{Recall}=95.02\%$ and $\text{avgIOU}=0.7$, \ie comparing manual to automatic annotation the first set has fewer annotations (5\% less Recall) and there exist some class misalignment (1.5\% less Precision).
To gain more insights on those differences, we visually examined the boxes obtained from the two sets of annotations and found out that the missing $1.5\%$ Precision can be mostly explained by class mistakes made by the human annotator during the labelling process, while \algoname{} cannot assign any wrong label by construction. The $5\%$ missing recall is instead due to situations like that depicted in \autoref{fig:dataset}(a-b), where the visible portion of an object (the bottom object \emph{Cls\_1}) is too small to allow the human annotator to recognize it.
Finally, the relatively low $avgIOU$ highlights a key difference between \algoname{} and a human annotator, the former always produces a box large enough to enclose the whole object as side effect of the re-projection of the virtual 3D box, while the latter usually encloses only the visible portion of the object (See the \autoref{fig:refinement_screens} (top-right) image where a virtual box is drawn also where the object is occluded). As a result, the manual and auto annotations does not always have matching shapes, especially in cluttered environment, as it can be observed in \autoref{fig:dataset} (a-b).
Nevertheless, as we will prove in the following paragraphs, the dataset labelled with \algoname{} can effectively be used to train and validate any machine learning based object detector obtaining performance comparable with a manually annotated dataset of the same size, while also enabling an effortless training with many more images so as to create quite very robust object detectors.   

\subsection{Object Detector Test}
\label{ssec:objDetTest}

\begin{figure*}
	\centering
	\begin{tabular}{ccc}
		\includegraphics[width=0.3\textwidth]{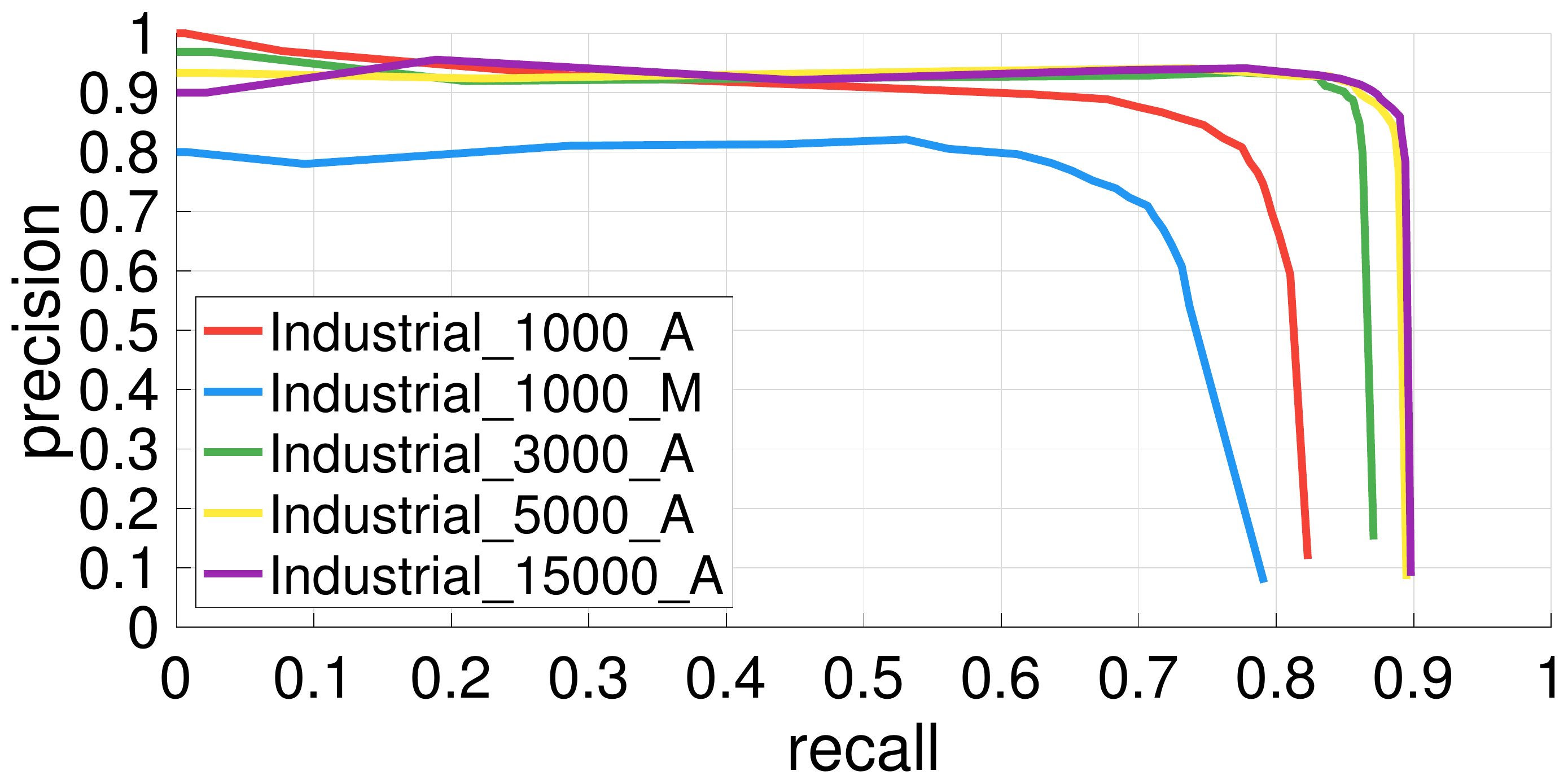} &
		\includegraphics[width=0.3\textwidth]{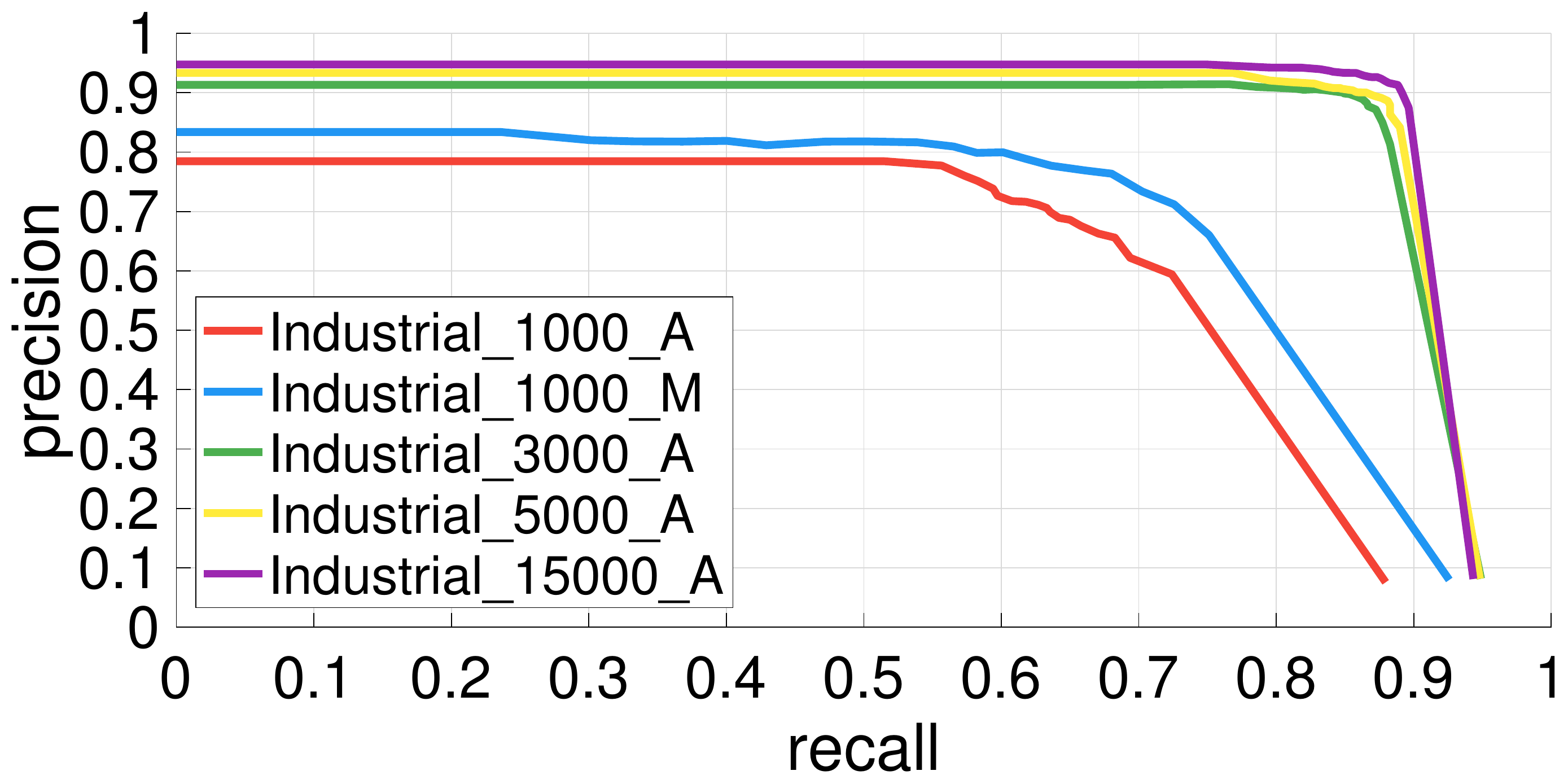} &
		\includegraphics[width=0.3\textwidth]{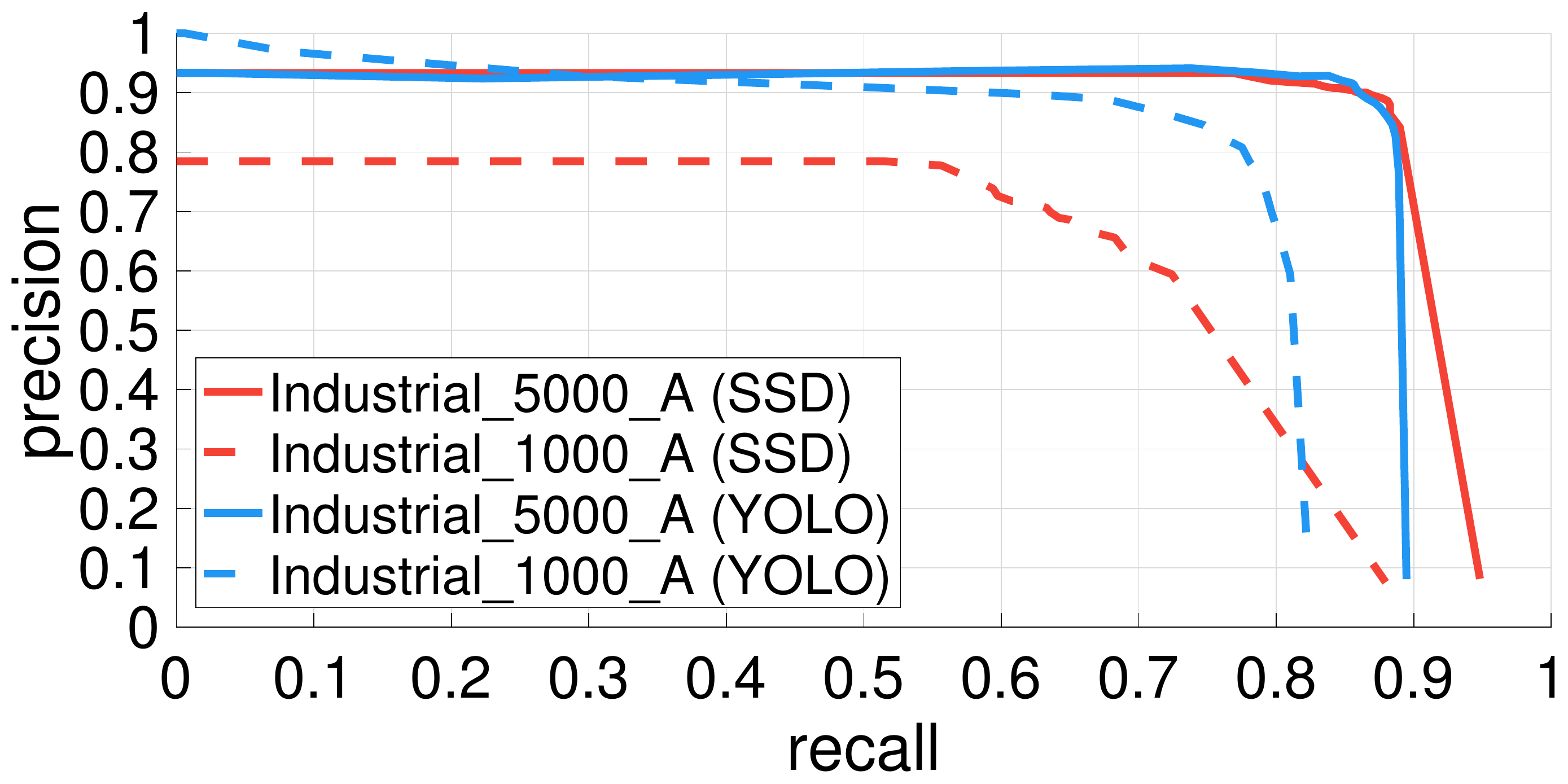} \\
		\textbf{(a)} YOLO & \textbf{(b)} SSD & \textbf{(c)} Comparison \\
	\end{tabular}
	\caption{Precision/Recall curves for the two type of detector trained on different subsets of the \elettrodataset{} dataset. (a) and (b) report the results for YOLO and SSD respectively; (c) instead displays a comparison between them.}
	\label{fig:industrial_results}
	
\end{figure*}

\begin{figure*}
	\centering
	\begin{tabular}{ccc}
		\includegraphics[width=0.3\textwidth]{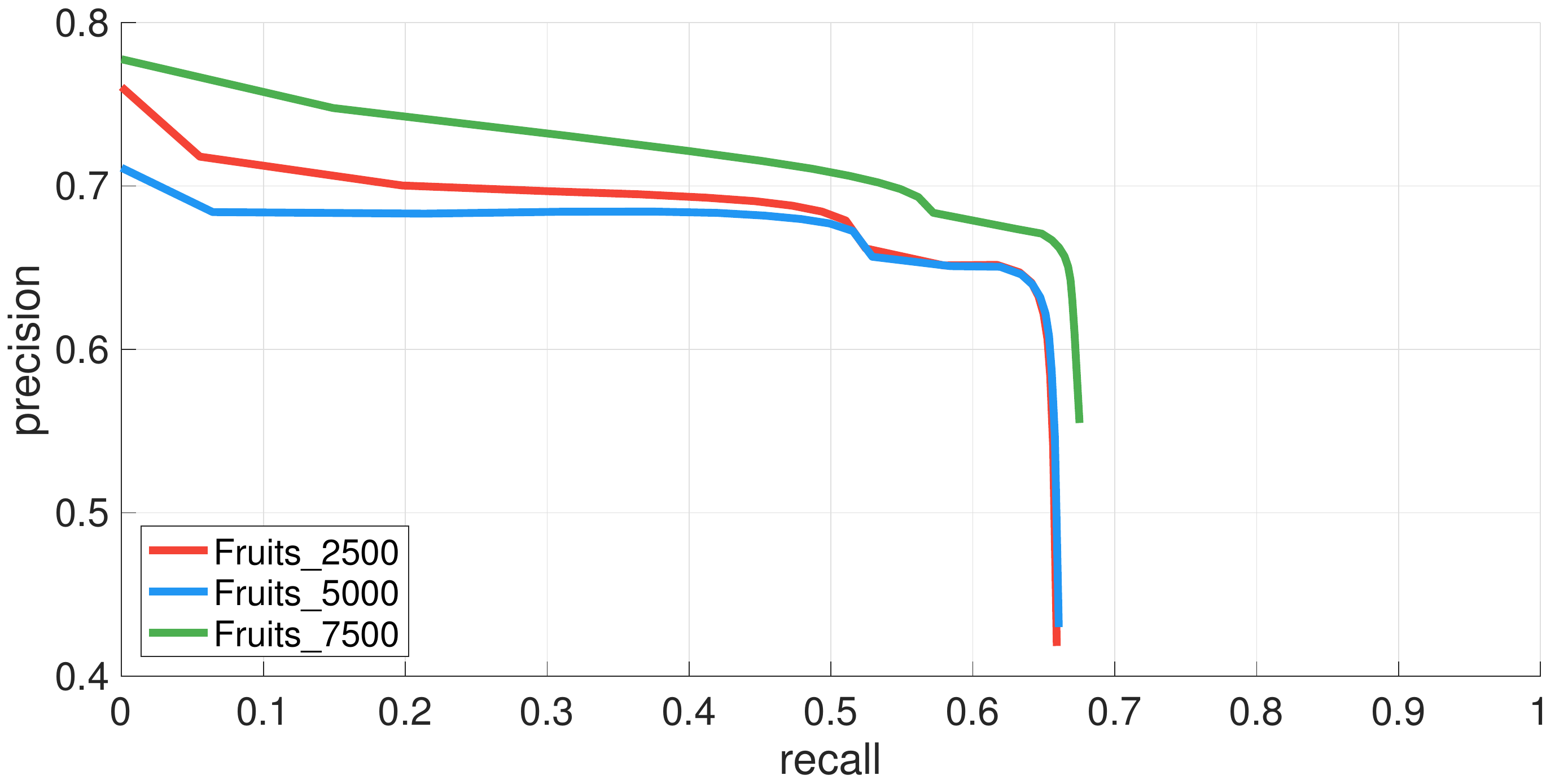} &
		\includegraphics[width=0.3\textwidth]{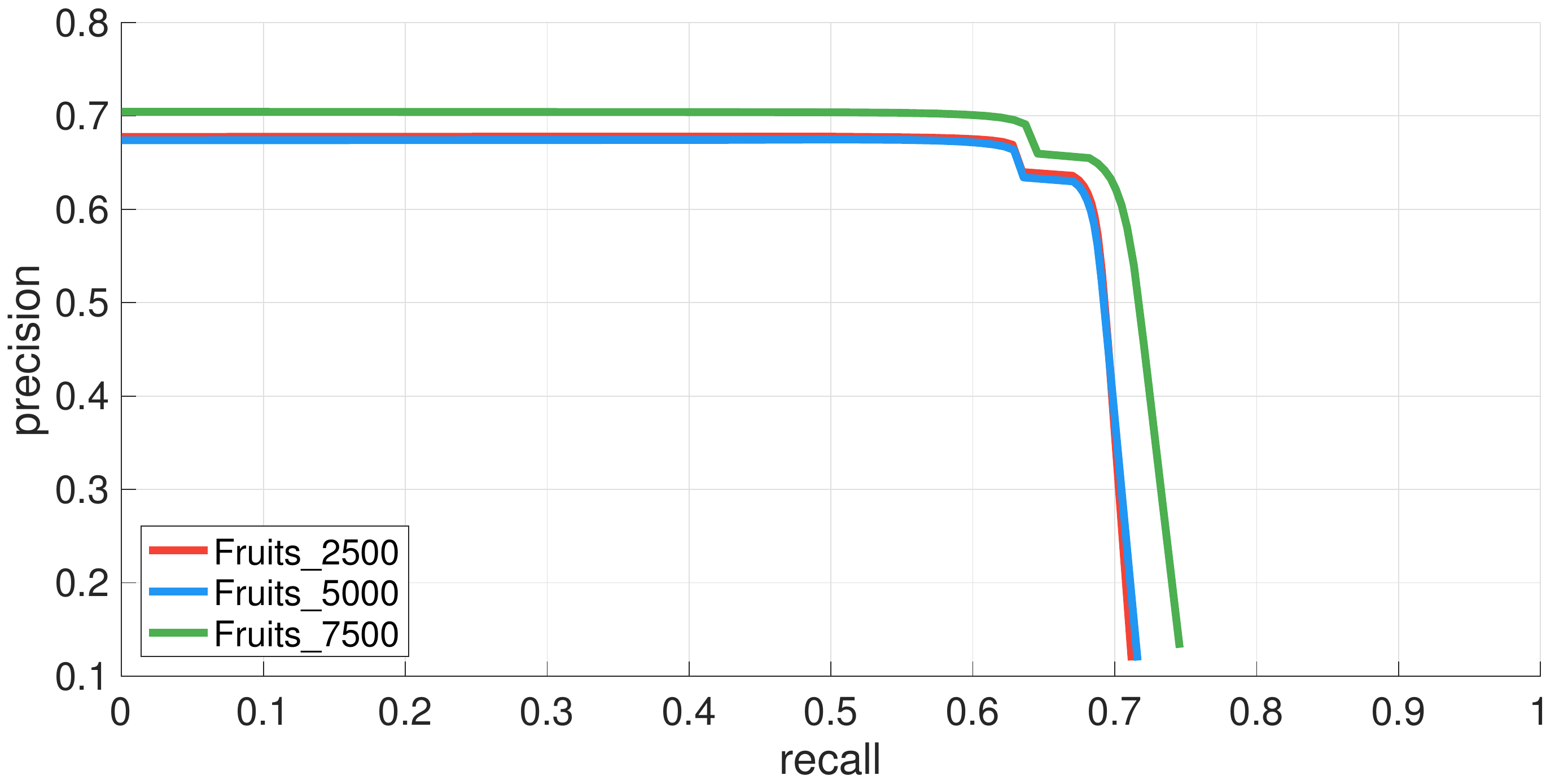} &
		\includegraphics[width=0.3\textwidth]{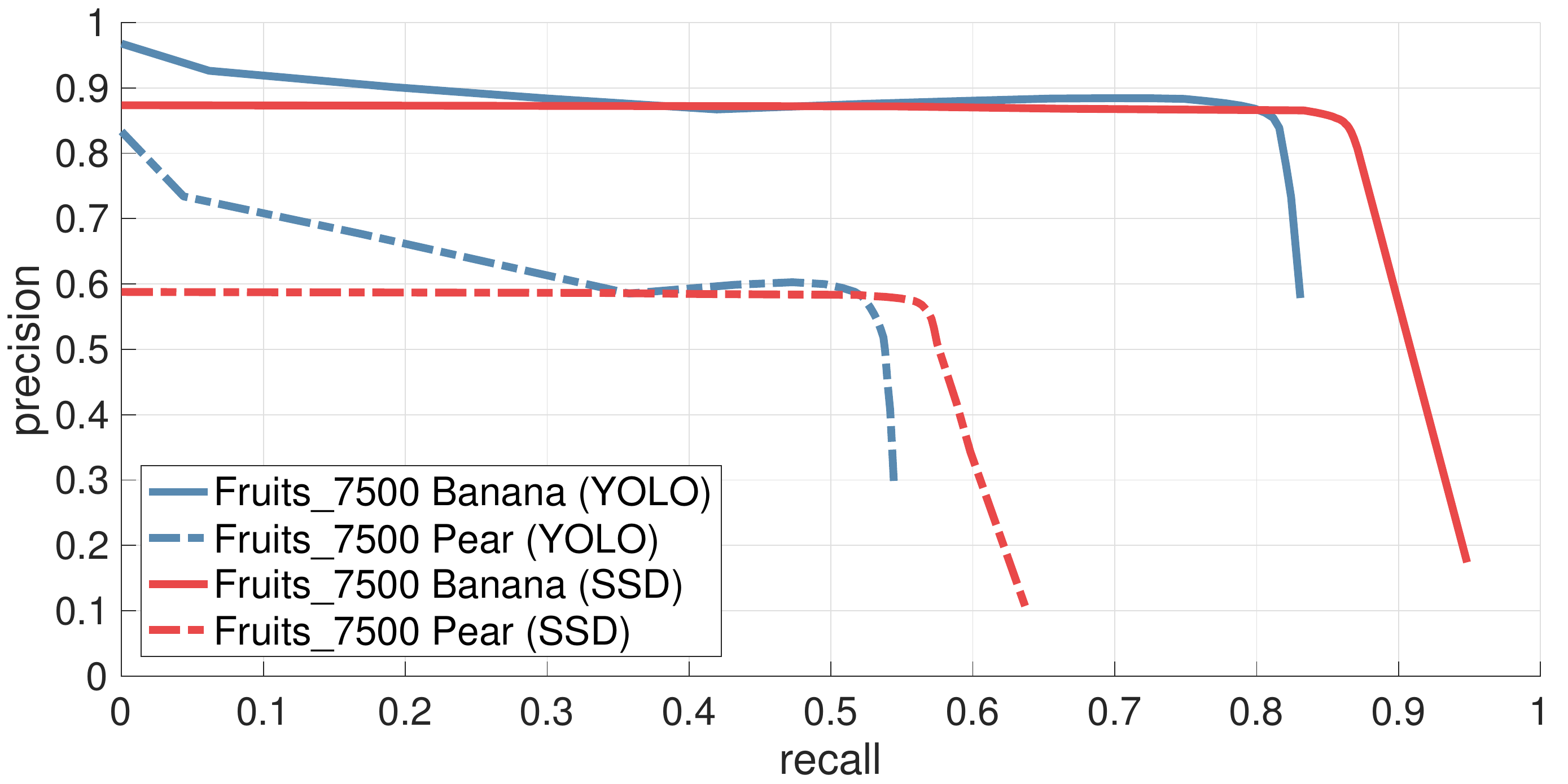}
		\\
		\textbf{(a)} YOLO & \textbf{(b)} SSD & \textbf{(c)} Class specific Comparison \\
	\end{tabular}
	\caption{Precision/Recall curves for the two type of detector trained on different subsets of the \fruitdataset{} dataset. (a) and (b) report the results for YOLO and SSD respectively; (c) instead displays a comparison between SSD and YOLO over the best class (\textit{Banana}) and the worst class (\textit{Pear}) of the \fruitdataset{} dataset.}
	\label{fig:fruitResult}
	
\end{figure*}

\begin{table}[t]
	\centering
	\resizebox{\columnwidth}{!}{%
	\begin{tabular}{|c|c|c|c|c|}
		\hline
		& \multicolumn{2}{c|}{\textbf{\emph{mAP}} (\text{th}=0.5)}& \multicolumn{2}{c|}{\textbf{\emph{avgIOU}}}\\
		\hline
		\textbf{Training set}& YOLO&SSD& YOLO&SSD\\
		\hline
		\elettrodataset{}\_1000\_M&0.589&0.619&0.7479&\textbf{0.795}\\
		\elettrodataset{}\_1000\_A&0.731&0.562&0.719&0.728\\
		\elettrodataset{}\_3000\_A&0.799&0.809&0.713&0.720\\
		\elettrodataset{}\_5000\_A&0.828&0.831&0.705&0.729\\
		\elettrodataset{}\_15000\_A&0.834&\textbf{0.851}&0.709&0.732\\
		\hline
	\end{tabular}%
	}
	\captionsetup{size=small,skip=0.333\baselineskip}
	\small
	\caption{Mean average precision (\emph{mAP}) and average intersection over union (\emph{avgIOU}) on the \elettrodataset{}\_Test+ for YOLO and SSD trained using 5 different training sets with increasing number of images. Best result  highlighted in bold.}
	\label{tab:elettroresult}
	
\end{table}

\begin{table}[t]
	\centering
	\resizebox{\columnwidth}{!}{%
	\begin{tabular}{|c|c|c|c|c|c|c|}
		\hline
		& \multicolumn{2}{c|}{\textbf{\emph{mAP}} (\text{th}=0.5)}& \multicolumn{2}{c|}{\textbf{\emph{mAP}} (\text{th}=0.3)}
        & \multicolumn{2}{c|}{\textbf{\emph{avgIOU}}}\\
		\hline
		\textbf{Training set}& YOLO&SSD& YOLO&SSD& YOLO&SSD\\
		\hline
		\fruitdataset{}\_2500\_A&0.438&0.468&0.895&0.888&0.710&0.744 \\
		\fruitdataset{}\_5000\_A&0.440&0.465&0.894&0.889&0.6818&0.749 \\
		\fruitdataset{}\_7500\_A&0.469&\textbf{0.504}&0.902&\textbf{0.904}&0.734&\textbf{0.756} \\
		\hline
	\end{tabular}%
}
	\captionsetup{size=small,skip=0.333\baselineskip}
	\small
	\caption{Mean average precision (\emph{mAP}), at two different $IOU_{th}$, and average intersection over union (\emph{avgIOU}) on \fruitdataset{}\_Test for YOLO and SSD trained using 3 different training sets with increasing number of images. "A" suffix marks training sets with annotations produced by \algoname{}, best result highlighted in bold.}
	\label{tab:fruitresult}
	
\end{table}

\begin{figure}
	\centering
		\includegraphics[width=1\columnwidth]{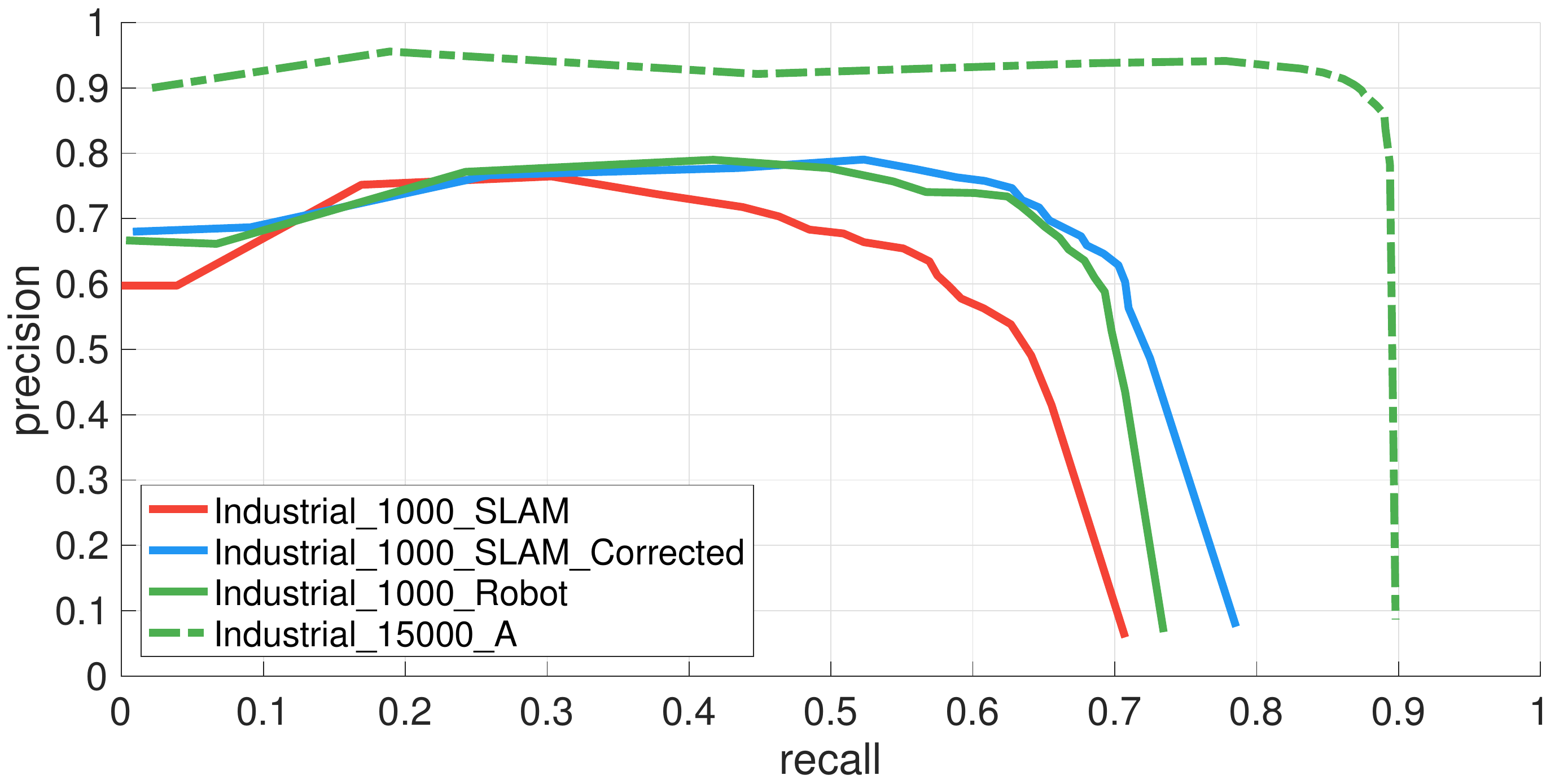}
	\caption{Precision/Recall curves for the YOLO detector trained on a subset of the  \elettrodataset{} dataset where the camera is also tracked with the \textit{Orb-Slam-v2} algorithm\cite{mur2017orb}.}
	
	\label{fig:slam_results}
	
\end{figure}

Once assessed that annotations obtained by \algoname{} are comparable with manual ones, we test how effective is an object detector system trained on them. We choose as detectors YOLO \cite{redmon2016yolo9000} and SSD \cite{huang2016speed}, using for both the original author's implementation and public pre-trained networks as initial weights.
The first tests concern the \elettrodataset{} dataset where we defined a test bench of 126 images randomly picked from \elettrodataset{}\_Test  plus 20 external smartphone pictures, all carefully manually annotated. We will call \elettrodataset{}\_Test+ this dataset.
Thanks to the fast labelling obtained by \algoname{} we were able to actually measure the performance boost  related to the training set size using four different sets with increasing number of images, respectively \elettrodataset{}\_1000, 3000, 5000 and 15000. The samples for  \elettrodataset{}\_1000 have both automatic and manual annotations, as to test if and how much the final detection performances change according to the label used. 

Given the five different training sets, we trained both detectors on them for 100000 steps with \texttt{batch\_size=24} using the default hyperparameters recommended by the authors. The results are ten slightly different detectors that we tested on \elettrodataset{}\_Test+, we reported in \autoref{fig:industrial_results} the obtained precision/recall curves and in \autoref{tab:elettroresult} the \emph{mAP} and \emph{avgIOU}.
As expected, for both detectors the performance increases proportionally to the size of the training set used, \eg +0.23 mAP gain between SSD trained on 1000 images and the best performing one trained on 15000; vouching the need for a method to ease and speed up the creation of huge training dataset.
Inspecting the \emph{avgIOU} obtained by the detectors we can see how the best performing methods are, unsurprisingly, the two trained on the manually annotated \elettrodataset{}\_1000\_M. It is due to the fact the the testing images are manually annotated, so a manual label is better suited for an IOU score.


We repeated similar experiments on the \fruitdataset{} dataset: we annotated all the eight sequences using \algoname{}, then we produced three different training and validation sets with increasing number of samples and sequences, respectively \fruitdataset{}\_2500 (2 sequences), 5000 (4 sequences) and 7500 (6 sequences).
As stated above, the remaining two sequences are used as the test bench for the detector creating a manual labelled test set of 1000 images (refereed as \fruitdataset{}\_Test).
We tested the six different resulting detectors on \fruitdataset{}\_Test and report the result in \autoref{fig:fruitResult} and in  \autoref{tab:fruitresult}. Once again all the performance indexes increase alongside  with the size of the training set, with best absolute performance obtained by SDD using the \fruitdataset{}\_7500 dataset.
 \autoref{fig:fruitResult}(c) reports an intra-detector comparison between YOLO and SSD on the best and the worst class. 


\subsection{Comparision between visual SLAM and robot-based camera tracking}\label{sec:slam_comparison}

Aiming at the comparison between the ARS approach and what can be achieved through different kinds of camera trackers, the state-of-the-art monocular SLAM described in \cite{mur2017orb} has been exploited to track the camera on the acquired datasets, and the results are compared with the ones obtained through robot based camera tracking previously described.
To this end, we use all the original Industrial training sequences to feed the visual SLAM algorithm, obtaining in this way a new version of the frame set in which the camera poses are estimated by tracking  the visual features instead of measuring it through the robot.

Unfortunately, state of the art SLAM system featuring a pose optimization graph, can produce camera poses only for a subset of frames (\ie the keyframes) much smaller than the original set.

In this specific case, the experimental compression ratio is about $1:30$, which results in reducing the dataset dimension to about 1000 images (starting from the 36000 of the Industrial dataset). We refer with Industrial 1000 SLAM to the direct outcome of the visual SLAM algorithm, while with Industrial 1000 Robot we refer to the corresponding frames in which the camera pose is estimated through the robot kinematics. 
Moreover, due to the known scaling factor problem of a monocular SLAM, the estimated camera poses may be slightly different from the real one, generating a non-perfect match with the original objects and, as a consequence, misaligned annotations.

However, by using ARS we manually corrected them with the procedure described in \autoref{sec:pose_refinement} producing an additional training set called Industrial 1000 SLAM Corrected. In Figure 8 we plot precision-recall curves obtained by YOLO trained on the three training sets. The performance of original SLAM procedure is worse than the corrected counterpart that is indeed comparable to the Robot version.

\begin{figure}
	\centering
		\includegraphics[width=1\columnwidth]{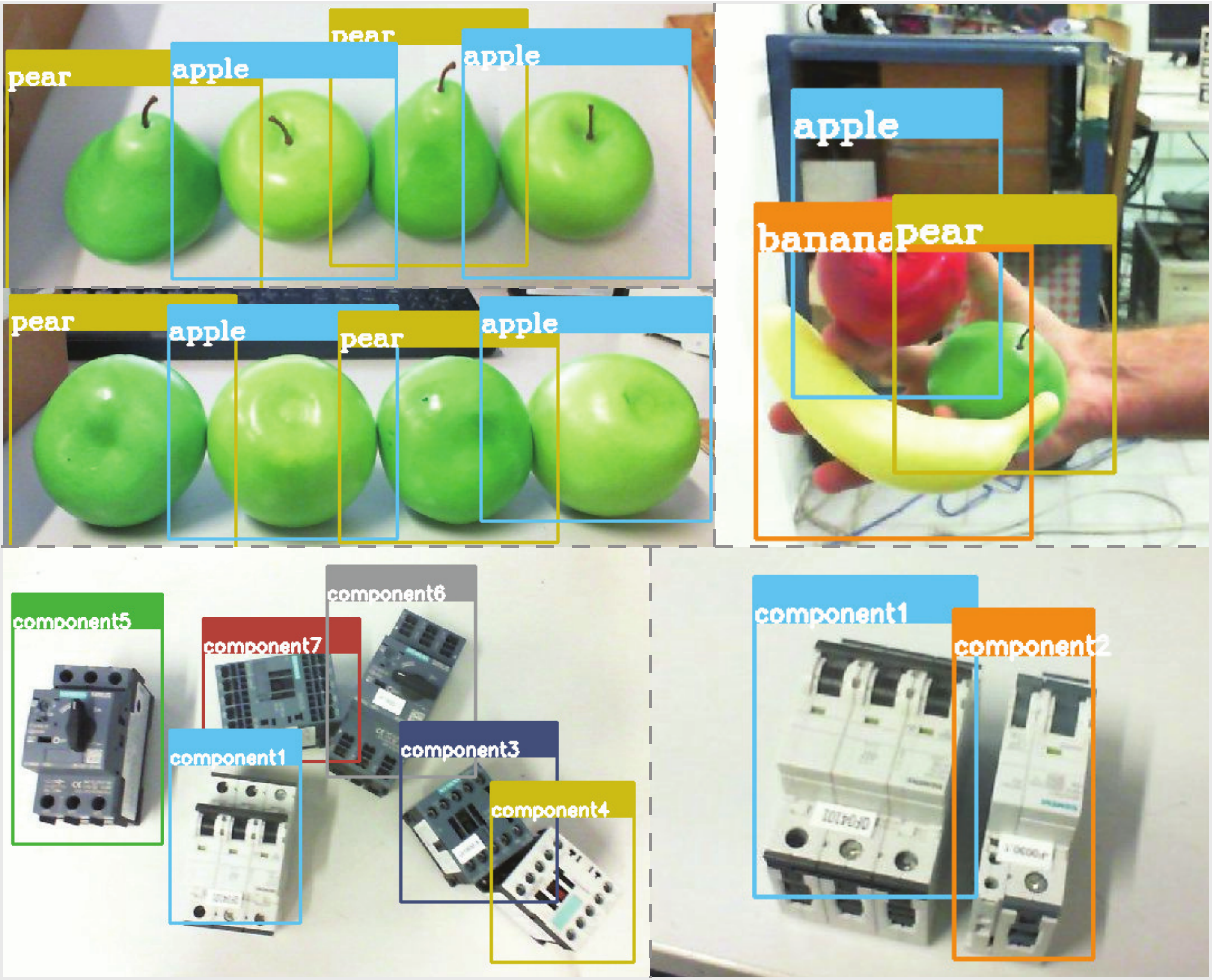}
	\caption{Detectors output (all correct) computed on images acquired with a usb webcam. The detector is able to distinguish \emph{apple}s and \emph{pear}s only looking at their lower side. }
	\label{fig:live_detections}
	
\end{figure}

\begin{figure*}
	\centering
		\includegraphics[width=0.9\textwidth]{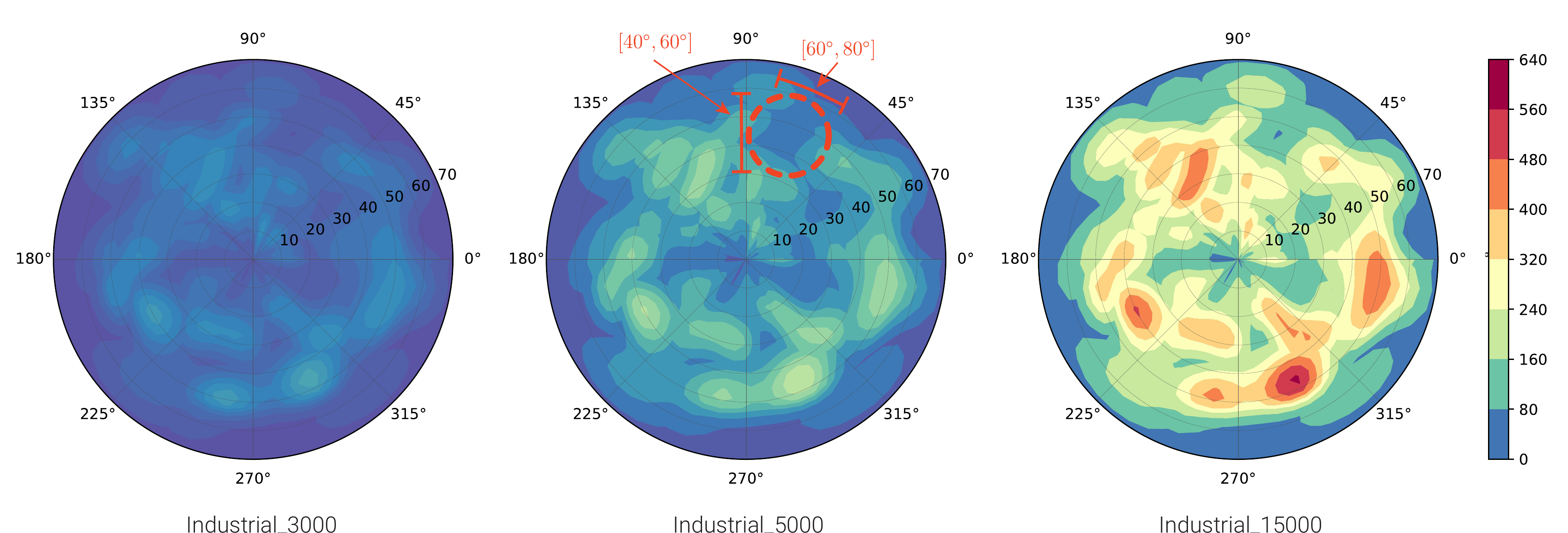}
	\caption{The \histoname{} computed for object of class 0 for the \elettrodataset{}\_3000/5000/15000 datasets. The ColorBar maps the color with the number of views voting for corresponding polar bin. }
	\label{fig:viewpoint_coverage}
\end{figure*}

\begin{figure*}
	\centering
	\begin{tabular}{cc}
		\includegraphics[width=0.6\textwidth]{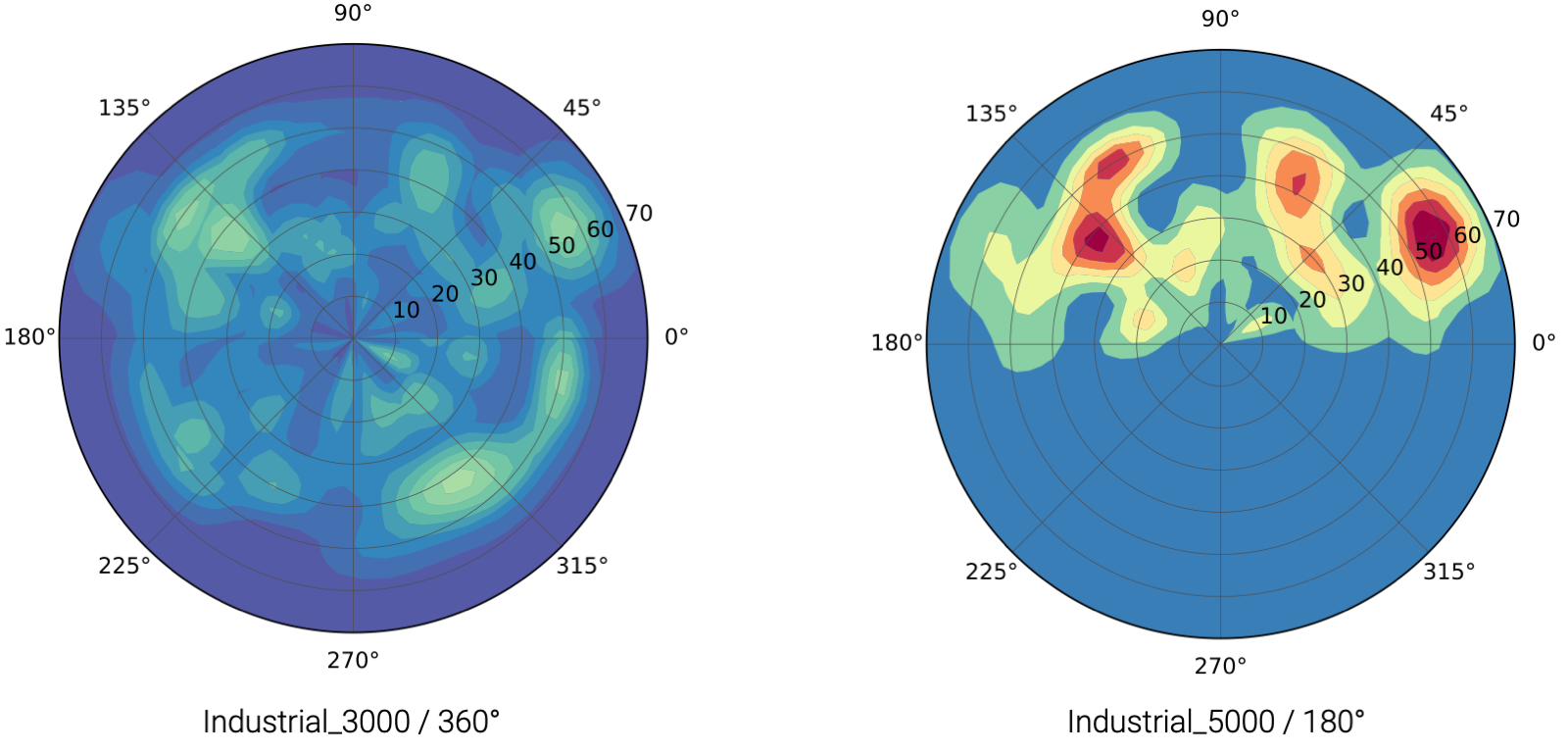} &
		
		\includegraphics[width=0.25\textwidth]{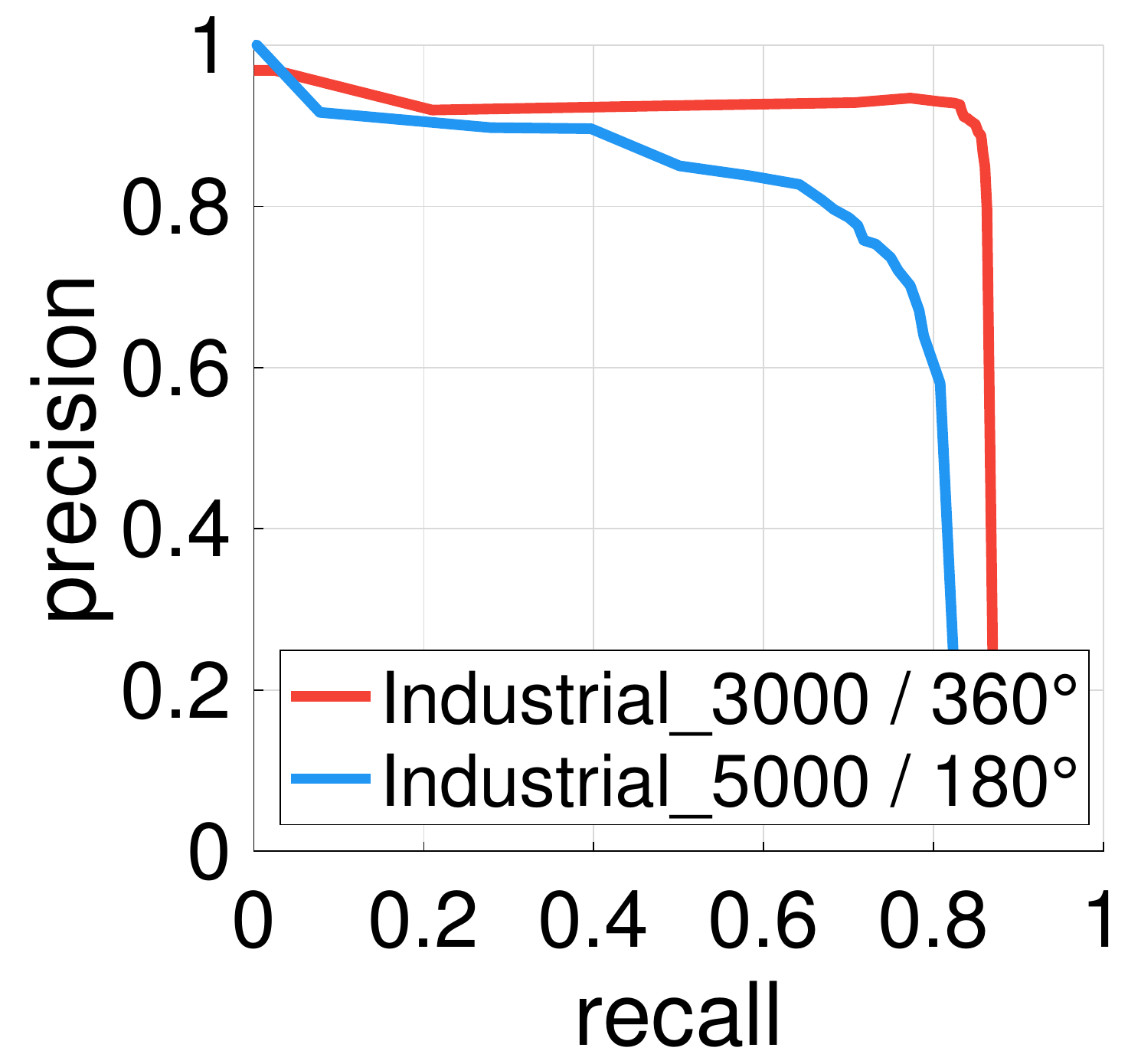} 
		\\  \textbf{(a)} & \textbf{(b)} 
	\end{tabular}
	\caption{(a) Shows the VC over  \elettrodataset{}\_3000 and \elettrodataset{}\_5000 with the latter filtered removing half the viewpoints. (b) Depicts the performance of an object detector trained on the two dataset, highlighting the importance of a full coverage at training time.}
	\label{fig:viewpoint}
	
\end{figure*}

\subsection{\histoname{}}
\label{ssec:histo}

All the results reported so far show that \algoname{} is fast and effective for the dataset creation, however, it has one additional useful side effect: for each image in the dataset we know the position of the camera with respect to each object in the scene. Thus, for each virtual box $\NInst_j=\{^{0}\TT_j , s_j, c_j \}$ we can compute the position of the camera \textit{w.r.t.} that object at the $i_{\text{th}}$ frame $^{j}\TT_{cam_i}$. We can express the position of the camera in the object RF in polar coordinates $^{j}\mathbf{p}_{cam_i}=^{j}(r,\theta,\phi)_{cam_i}$ (\ie \emph{radial}, \emph{azimuthal}, \emph{polar}) and build a 2D histogram by aggregating $(\theta,\phi)$ into bins and counting the number of frames which contribute to that viewpoint. We dubbed this histogram the \emph{Viewpoints Coverage} (VC) of an object in the training dataset. \autoref{fig:viewpoint_coverage} shows as  heat map the histograms for object class $0$ on three of the \elettrodataset{} training sets presented above, with hotter colors corresponding to higher coverage, \ie to more frames acquired from that viewpoint. On the middle histogram of \autoref{fig:viewpoint_coverage} we highlighted with a dashed circle a region with low score, \ie a viewpoint poorly covered in the training set and thus a potential flaw in the final detector when watching the object from that vantage point in a test image. Therefore the VC representation may be used during the creation of the dataset to guide the user (or a Robot), \eg suggesting how to acquire new sequences carrying out better trajectories.
To highlight the importance of having as much object coverage as possible we define two dataset \elettrodataset{}\_3000/360$^{\circ}$ (featuring only 3000 images but covering all viewpoints) and \elettrodataset{}\_5000/180$^{\circ}$ (featuring 5000 images but only covering half of the possible viewpoints) considering only object class $0$, with \autoref{fig:viewpoint}(a) depicting the corresponding VCs. We use those as training set for  YOLO and report in \autoref{fig:viewpoint}(b) the resulting precision/recall curves for that object class. As expected, even featuring 2000 images more, \elettrodataset{}\_5000/180$^{\circ}$ perform much worse due to having seen only a limited set of object appearances.  In conclusion, with the Viewpoint Coverage analysis we demonstrate that methods like this are crucial to control the distribution of training data	, which is much more relevant than the size of an uncontrolled dataset. In addition, in case of ARS deployed in a robotic scenario, the VC metrics could guide the robot to perform optimal trajectories to maximize the coverage of the viewpoints autonomously, or, in case of impossible trajectories, inform the user that there is a need to rearrange objects in the scene.

\section{Conclusion and Future Work}

In this paper, we demonstrated how, by using robotic vision (i.e. robotics at the service of vision and vice versa), it is possible to create systems, such as the one here presented, that use the robot itself to learn. By exploiting the dexterity of the robot, a large number of viewpoints are automatically generated for each object (as depicted in \autoref{fig:viewpoint_coverage}), to allow the neural networks to generalize well distilling a knowledge of them. Moreover, the possibility to generate self-annotated images without human intervention allows to effortlessly collect countless environmental variations (such as light or working table color changes) thus generating robust visual perception even in unstructured environments.

By the proposed approach, two novel datasets are effortlessly created, one on electromechanical components (industrial scenario) and one on fruits (daily-living scenario). From these datasets, two state-of-the-art object detectors based on convolutional neural networks, such as YOLO and SSD, are trained robustly.  The proposed approach based on ARS allows to annotate 9 sequences of about 35000 frames in less than one hour, that compared to conventional manual annotation of 1000 frames that takes us slightly more than 10 hours, results in a gain factor of about 450 considering both the time saved and the dataset dimension. From the point of view of performance in the object detection, both the precision and recall is increased by about 15\% with respect to manual labelling. The proposed approach allows to embed into robots novel and more performing perception and learning capabilities at the expense of a very limited human intervention. All the software generated to implement the proposed approach is available as a ROS package in a public repository alongside with the novel annotated datasets \footnote{\url{https://github.com/m4nh/ars}}.

As a future extension of the proposed method, we will exploit the knowledge of the 6-DOF pose of objects with respect to the camera provided by the robot in each viewpoint to train more complex systems than a simple 2D detector. To this end, the approaches adopted to estimate the 3D position and orientation of objects from single 2D images reported in literature, see for example \cite{Kehl_2017_ICCV,rad2017bb8,sundermeyer2018implicit}, will be trained with the output provided by the ARS pipeline. This approach will enable the realization of a fully automated self-learning BinPicking system. 

Another \algoname{} extension that we plan to explore concerns the use of out of the box augmented reality toolkit offered by nowadays  mobile platforms
(\eg ARKit for iOS and ARCore for Android) in order to track the camera pose while acquiring video sequences. This extension would allow quick and easy creation of a training dataset by an off-the-shelf mobile device such as a smartphone or a tablet.



\bibliographystyle{spmpsci}  
\bibliography{biblio}

\begin{IEEEbiography}
[{\includegraphics[width=1in,height=1.25in,clip,keepaspectratio]{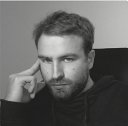}}]
{Daniele De Gregorio}
received the B.Sc. and M.Sc. degrees from the University of L'Aquila,
Italy, in 2008 and 2012, respectively, and the Ph.D. degree in software engineering from the University of Bologna, Italy, in 2018. He is currently a Post-Doctoral Researcher with the University of Bologna. His research interests include robotics, computer vision, and deep learning. 
\end{IEEEbiography}
\vspace{-1cm}
\begin{IEEEbiography}
[{\includegraphics[width=1in,height=1.25in,clip,keepaspectratio]{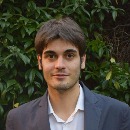}}]
{Alessio Tonioni}
Received his PhD degree in Computer Science and Engineering from University of Bologna in 2019.
Currently, he is a Post-doc researcher at Department of Computer Science and Engineering, University of Bologna. His research interest concerns machine learning for depth estimation and object detection.
\end{IEEEbiography}
\vspace{-1cm}
\begin{IEEEbiography}
[{\includegraphics[width=1in,height=1.25in,clip,keepaspectratio]{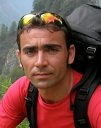}}]
{Gianluca Palli}
received the Laurea and
Ph.D. degrees in automation engineering from the
University of Bologna, Bologna, Italy, in 2003 and
2007, respectively.
He is currently an Associate Professor with
the University of Bologna. He is an author
or a co-author over 80 scientific papers presented
at conferences or published in journals. 
\end{IEEEbiography}
\vspace{-1cm}
\begin{IEEEbiography}
[{\includegraphics[width=1in,height=1.25in,clip,keepaspectratio]{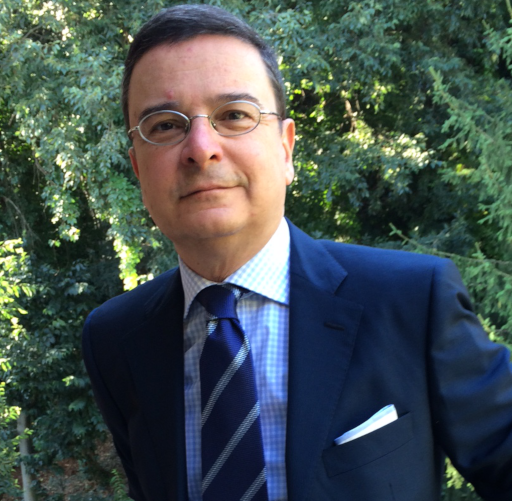}}]
{Luigi Di Stefano}
received the PhD degree in electronic engineering and computer science from the University of Bologna in 1994. He is currently a full professor at the Department of Computer Science and Engineering, University of Bologna, where he founded and leads the Computer Vision Laboratory (CVLab). His research interests include image processing, computer vision and machine/deep learning. He is the author of more than 150
papers and several patents. He has been scientific consultant for major companies in the fields of computer vision and machine learning. 
He is a member of the IEEE Computer Society and the IAPR-IC.
\end{IEEEbiography}

\end{document}